\documentclass[twoside]{article}

\usepackage[accepted]{aistats2024}
\usepackage[round]{natbib}
\usepackage{hyperref}       % hyperlinks
\usepackage{url}            % simple URL typesetting
\usepackage{booktabs}       % professional-quality tables
\usepackage{amsfonts}       % blackboard math symbols
\usepackage{nicefrac}       % compact symbols for 1/2, etc.
\usepackage{microtype}      % microtypography
\usepackage{xcolor}         % colors

\usepackage{bm}
\usepackage{threeparttable}
\usepackage{multirow}
\usepackage{subcaption}
\usepackage{graphicx}
\usepackage{amsmath}
\usepackage{amsthm}

\newtheorem{theorem}{Theorem}
\newtheorem{proposition}[theorem]{Proposition}

\newtheorem{definition}[theorem]{Definition}
\newtheorem{assumption}[theorem]{Assumption}

\usepackage{algorithm}
\usepackage{algorithmic}
\begin{document}

% If your paper is accepted and the title of your paper is very long,
% the style will print as headings an error message. Use the following
% command to supply a shorter title of your paper so that it can be
% used as headings.
%
\runningtitle{Invariant Aggregator for Defending against Federated Backdoor Attacks}

% If your paper is accepted and the number of authors is large, the
% style will print as headings an error message. Use the following
% command to supply a shorter version of the authors names so that
% they can be used as headings (for example, use only the surnames)
%
\runningauthor{Xiaoyang Wang, Dimitrios Dimitriadis, Sanmi Koyejo, Shruti Tople}

\twocolumn[

\aistatstitle{Invariant Aggregator for Defending against \\ Federated Backdoor Attacks}

\aistatsauthor{Xiaoyang Wang$^*$ \And Dimitrios Dimitriadis$^\dagger$}

\aistatsaddress{University of Illinois Urbana-Champaign \And } 
\aistatsemail{\href{mailto:xw28@illinois.edu}{xw28@illinois.edu}  \And \href{mailto:ddimitriadis@gmail.com}{ddimitriadis@gmail.com}} 

\aistatsauthor{Sanmi Koyejo$^\dagger$ \And Shruti Tople$^\dagger$}
\aistatsaddress{Stanford University \And Azure Research} 
\aistatsemail{\href{mailto:sanmi@stanford.edu}{sanmi@stanford.edu} \And \href{mailto:Shruti.Tople@microsoft.com}{Shruti.Tople@microsoft.com}} 
]

\begin{abstract}
\vspace{-0.2in}
Federated learning enables training high-utility models across several clients without directly sharing their private data. As a downside, the federated setting makes the model vulnerable to various adversarial attacks in the presence of malicious clients. Despite the theoretical and empirical success in defending against attacks that aim to degrade models' utility, defense against backdoor attacks that increase model accuracy on backdoor samples exclusively without hurting the utility on other samples remains challenging. To this end, we first analyze the failure modes of existing defenses over a flat loss landscape, which is common for well-designed neural networks such as Resnet \citep{He2015DeepRL} but is often overlooked by previous works. Then, we propose an invariant aggregator that redirects the aggregated update to invariant directions that are generally useful via selectively masking out the update elements that favor few and possibly malicious clients. Theoretical results suggest that our approach provably mitigates backdoor attacks and remains effective over flat loss landscapes. Empirical results on three datasets with different modalities and varying numbers of clients further demonstrate that our approach mitigates a broad class of backdoor attacks with a negligible cost on the model utility.
\end{abstract}

% \vspace{-0.2in}
\section{Introduction}
% \vspace{-0.05in}
Federated learning enables multiple distrusting clients to jointly train a machine learning model without sharing their private data directly. However, a rising concern in this setting is the ability of potentially malicious clients to perpetrate backdoor attacks and control model predictions using a backdoor trigger \citep{Liu2018TrojaningAO, pmlr-v108-bagdasaryan20a}. To this end, it has been argued that conducting backdoor attacks in a federated learning setup is practical \citep{9833647} and can be effective \citep{NEURIPS2020_b8ffa41d}. The impact of such attacks is quite severe in many mission-critical federated learning applications. For example, anomaly detection is a common federated learning task where multiple parties (e.g., banks or email users) collaboratively train a model that detects fraud or phishing emails. Backdoor attacks allow the adversary to circumvent these detections successfully.

% \vspace{-0.15in}

\paragraph{Motivating Setting.} To better develop a defense approach, we first analyze the vulnerability of federated learning systems against backdoor attacks over a flat loss landscape. A flat loss landscape is considered an essential factor in the empirical success of neural network optimization \citep{Li2017VisualizingTL, Sun2020TheGL}. Although neural networks are non-convex in general and may have complicated landscapes, recent works \citep{Li2017VisualizingTL, santurkar2018does} suggest that improved neural network architecture design such as the Resnet with skip connections \citep{He2015DeepRL} can significantly flatten the loss landscape and ease the optimization. As a downside, a flat loss landscape may allow manipulation of model parameters without hurting the utility on benign samples, which is precisely the phenomenon that backdoor adversaries easily exploit. A key insight is that backdoor attacks over flat loss landscapes can succeed without incurring significant differences between benign and malicious client updates due to the diminished gradient magnitudes from benign clients. We further show that this phenomenon, combined with other factors, such as the stochastic nature of the update, can help backdoor adversaries circumvent existing defenses. Our analysis also broadly includes data-centric approaches such as the edge-case attack \citep{NEURIPS2020_b8ffa41d} and the trigger inversion defense \citep{Wang2019NeuralCI, zhang2023flip}.

% \vspace{-0.05in}

\textbf{Our methodology.} To avoid the failure modes of existing defenses over flat loss landscapes, we propose an invariant aggregator to defend against federated backdoor attacks under a minority adversary setting \citep{9833647}. Our defense examines each dimension of (pseudo-)gradients\footnote{We overload "gradient" to indicate any local model update communicated to the server in the federated setting, e.g., updates could be pseudo-gradients computed as differences between model updates after several local steps.} to avoid overlooking any backdoor attacks that only manipulate a few elements without incurring much difference on gradient vectors. For each dimension, we enforce the aggregated update points to invariant directions that are generally useful for most clients instead of favoring a few and possibly malicious clients. As a result, our defense remains effective with flat loss landscapes where the magnitudes of benign gradients can be small.

% \vspace{-0.05in}

\textbf{Our approach.} We consider the gradient sign (e.g., positive, negative, or zero) as a magnitude-agnostic indicator of benefit. Two clients having a \textit{consistent sign} implies that going along the direction pointed by the gradient can benefit both clients and vice versa. Following this intuition, we employ an AND-mask \citep{parascandolo2021learning} to set the gradient dimension with sign consistency below a given threshold to zero, masking out gradient elements that benefit a few clients. However, this alone is insufficient: the malicious clients can still use outliers to mislead the aggregation result even if the sign consistency is high. To address this issue, we further complement AND-mask with the trimmed-mean estimator \citep{10.1007/978-3-030-46147-8_13, 10.1214/20-AOS1961} as a means to remove the outliers. We theoretically show that the combination of AND-mask and trimmed-mean estimator is necessary and sufficient for mitigating backdoor attacks.

% Summary of experiments.
Our empirical evaluation employs a broad class of backdoor attacks, as detailed in Section \ref{section: setup}, to test our defense. Empirical results on tabular (phishing emails), visual (CIFAR-10) \citep{Krizhevsky2009LearningML, pmlr-v54-mcmahan17a}, and text (Twitter) \citep{Caldas2018LEAFAB} datasets demonstrate that our method is effective in defending against backdoor attacks without degrading utility as compared to prior works. On average, our approach decreases the backdoor attack success rate by 61.6\% and only loses 1.2\% accuracy on benign samples compared to the standard FedAvg aggregator \citep{pmlr-v54-mcmahan17a}.

% Main contribution
\textbf{Contributions.} Our contributions are as follows:
\begin{itemize}
    \item We analyze the failure modes of multiple prominent defenses against federated backdoor attacks over a flat loss landscape.
    \item We develop a combination of defenses using an AND-mask and the trimmed-mean estimator against the backdoor attack by focusing on the dimension-wise invariant gradient directions.
    \item We theoretically analyze our strategy and demonstrate that a combination of an AND-mask and the trimmed-mean estimator is necessary and sufficient for mitigating backdoor attacks. 
    \item We empirically evaluate our method on three datasets with varying modality, trigger patterns, model architecture, and client numbers, as well as comparing the performance to existing defenses.
\end{itemize}

\section{Related Work}

\textbf{Backdoor Attack.} Common backdoor attacks aim at misleading the model predictions using a trigger \citep{Liu2018TrojaningAO}. The trigger can be digital \citep{pmlr-v108-bagdasaryan20a}, physical \citep{Wenger2021BackdoorAA}, semantic \citep{NEURIPS2020_b8ffa41d}, or invisible \citep{Li2021InvisibleBA}. Recent works extended backdoor attacks to the federated learning setting and proposed effective improvements such as gradient scaling \citep{pmlr-v108-bagdasaryan20a} or generating edge-case backdoor samples \citep{NEURIPS2020_b8ffa41d}. The edge-case backdoor attack shows that using backdoor samples with low probability density on benign clients (i.e., unlikely samples w.r.t. the training distribution) is hard to detect and defend in the federated learning setting.

\textbf{Centralized Defense.}
There is a line of work proposing centralized defenses against backdoor attacks where the main aim is either detecting the backdoor samples \citep{NEURIPS2018_280cf18b} or purifying the model parameters that are poisoned \citep{li2021anti}. However, applying such centralized defense to federated learning systems is infeasible in practice due to limited client data access in many implementations.

\textbf{Federated Defenses.}
Recent works have attempted to defend against backdoor attacks in federated learning systems. \cite{Sun2019CanYR} shows that weak differential-private (weak-dp) federated averaging can mitigate the backdoor attack. However, the weak-dp defense is circumvented by the improved edge-case federated backdoor attack \citep{NEURIPS2020_b8ffa41d}. \cite{Nguyen2021FLAMETB} suggests that the vector-wise cosine similarity can help detect malicious clients performing backdoor attacks. The vector-wise cosine similarity is insufficient when the backdoor attacks can succeed with few poisoned parameters, incurring little vector-wise difference \citep{wu2021adversarial}. Other defenses against untargeted poisoning attacks \citep{Blanchard2017MachineLW, 10.1007/978-3-030-46147-8_13} lack robustness against the backdoor attack. Sign-SGD with majority vote \citep{pmlr-v80-bernstein18a, Bernstein2019signSGDWM} is similar to our approach, but it always takes the majority direction instead of focusing on the invariant directions. Unlike existing works, our defense encourages the model to pursue invariant directions in the optimization procedure.

% \textbf{Certification.}
% Different from the above-discussed defenses, certification \citep{pmlr-v139-xie21a} aims at extinguishing backdoor samples within a neighborhood of a benign sample. A direct comparison between certification and our defense is not meaningful due to the different evaluation metrics. Certification considers the certification rate of benign samples as the metric, while our defense aims to reduce the backdoor samples' accuracy. However, it would be interesting to investigate whether the proposed defense can ease the certification of a model.

\section{Preliminaries}

\subsection{Notation} We assume a synchronous federated learning system, where $N$ clients collaboratively train an ML model $f: \mathcal{X} \rightarrow \mathcal{Y}$ with parameter $\bm{w} \in \mathbb{R}^\mathrm{d}$ coordinated by a server. An input to the model is a sample $\bm{x} \in \mathcal{X}$ with a label $y$. There are $N' < \frac{N}{2}$ adversarial clients aiming at corrupting the ML model during training \citep{9833647}. The $i^\mathrm{th}$, $i \in [1, ..., N]$, client has $n_i$ data samples, being benign for $i \in [1, ..., N - N']$ or being adversarial for $i \in [N - N' + 1, ..., N]$. The synchronous federated learning is conducted in $T$ rounds. In each round $t \in [1, ..., T]$, the server broadcasts a model parameterized by $\bm{w}_{t-1}$ to all the participating clients. We omit the subscript $t$ while focusing on a single round. Then, the $i^\mathrm{th}$ client optimizes $\bm{w}_{t-1}$ on their local data samples indexed by $j$ and reports the locally optimized $\bm{w}_{t, i}$ to the server. We define pseudo-gradient $\bm{g}_{t, i} = \bm{w}_{t - 1} - \bm{w}_{t, i}$ being the difference between the locally optimized model and the broadcasted model from the previous round. For simplicity, we often use the term ``gradient'' to refer to the pseudo-gradient. Once all gradients are uploaded, the server aggregates them and produces a new model with parameters $\bm{w}_t$ using the following rule: $\bm{w}_t = \bm{w}_{t-1} - \sum_{i=1}^{N}\frac{n_i}{\sum_{i=1}^{N}n_i} \bm{g}_{t, i}$. The goal of federated learning is to minimize a weighted risk function over the $N$ clients: $\mathcal{L}(\bm{w}) =  \sum_{i=1}^{N}\frac{n_i}{\sum_{i=1}^{N}n_i} \mathcal{L}_i(\bm{w}) = \sum_{i=1}^{N}\frac{n_i}{\sum_{i=1}^{N}n_i} \mathbb{E}_{\mathcal{D}_i}[\ell(f(x; \bm{w}), y)]$, where $\ell: \mathbb{R} \times \mathcal{Y} \rightarrow \mathbb{R}$ is a loss function. $\mathrm{sign}(\cdot)$ denotes an element-wise sign operator, $\odot$ denotes the Hadamard product operator, and $\mathrm{W}_1(\cdot, \cdot)$ denotes the Wasserstein-1 distance.

\subsection{Threat Model} The adversary generates a backdoor data sample $\bm{x}'$ by embedding a trigger in a benign data sample $\bm{x}$ and correlating the trigger with a label $y'$, which is different from the label $y$ of the benign data sample. We use $\mathcal{D}'$ to denote the distribution of backdoor data samples. Then, the malicious clients connect to the federated learning system and insert backdoor data samples into the training set. Since federated learning aims to minimize the risk over all clients' datasets, the model can entangle the backdoor signals while trying to minimize the risk over all clients. 

\subsection{Assumptions}
\label{section: assumptions}
Bounded heterogeneity is a common assumption in federated learning literature \citep{wang2021field}. Let $\bm{w}^*_i$ be a minimum in client $i$'s loss landscape. We assume the distance between the minimum of benign clients is bounded. Here, $\bm{w}^*_i$ is not necessarily a global minimum or a minimum of any global federated learning model but a parameter that a local model would converge to alone if client $i$ has a sufficient amount of data.
 
\begin{assumption} (Bounded heterogeneity)
\label{assumption: bounded heterogeneity}
    $\|\bm{w}^*_i - \bm{w}^*_j\| \leq \delta, \forall i \neq j, i \leq N - N', j \leq N - N'$.
\end{assumption}

Let $\mathcal{W}^*$ be a convex hull of $\{\bm{w}^*_i \mid i = 1, ..., N - N'\}$, we assume that malicious clients aim to converge to a model $\bm{w}'$ that is not in the convex hull $\mathcal{W}^*$ of benign clients' minima. However, we do not assume that all parameters in the convex hull $\mathcal{W}^*$ lead to zero backdoor success rate, especially since the convex hull may increase as the diameter $\delta$ of $\mathcal{W}^*$ increases. We empirically justify this separability assumption in Appendix \ref{appendix: experimental results}. Formally, this separability assumption is stated in the following.

% \vspace{-0.05in}
\begin{assumption} (Separable minimum)
\label{assumption: separable minimum}
    Let $\mathcal{W}^*$ be a convex hull with diameter $\delta$ of benign minima $\{\bm{w}^*_i \mid i = 1, ..., N - N'\}$ and $\bm{w}'$ be a minimum of malicious client, we have $\bm{w}' \notin \mathcal{W}^*$. 
\end{assumption}
% \vspace{-0.05in}

The estimated gradient often differs from the expected gradient in stochastic gradient descent. One of the most common models for estimated gradients is the additive noise model, which adds a noise term (e.g., Gaussian noise) to the expected gradient \citep{Wu2019OnTN}. For a given noise magnitude, the directional change of an estimated gradient may increase if its corresponding expected gradient shrinks. Formally, this noise assumption is stated in the following.

% \vspace{-0.05in}
\begin{assumption} (Noisy gradient estimation)
\label{assumption: noisy gradient}
    Let $\bm{g}_i$ be an estimated gradient vector from client $i$, we assume $\bm{g}_i = \mathbb{E}_{\mathcal{D}_i}[\bm{g}_i] + \epsilon_i$ where the noise term $\epsilon_i \sim \mathcal{N}(0, \bm{\sigma}_i)$ and $\mathcal{N}(0, \bm{\sigma}_i)$ is a Gaussian distribution with finite-norm covariance matrix $\bm{\sigma}_i$.
\end{assumption}

\section{Motivating Setting}
\vspace{-0.05in}
\label{section: motivating setting}

\begin{figure*}
  \centering
  \begin{subfigure}[b]{0.3\linewidth}
    \centering
    \includegraphics[width=\textwidth]{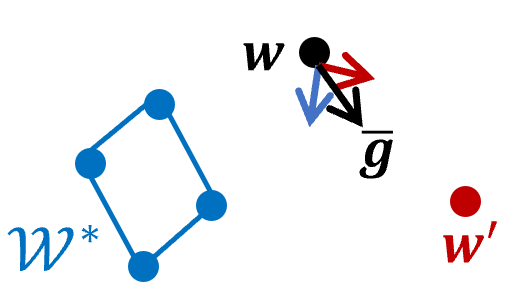}
    \caption{Overview\label{figure: minima}}
  \end{subfigure}
  \hfill
  \begin{subfigure}[b]{0.3\linewidth}
    \centering
    \includegraphics[width=0.75\textwidth]{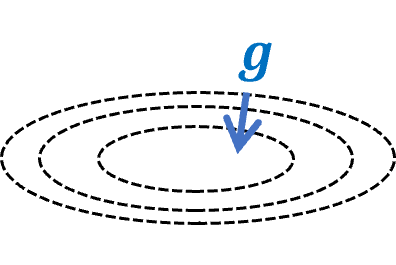}
    \caption{Landscape on a benign client\label{figure: benign region}}
  \end{subfigure}
  \hfill
  \begin{subfigure}[b]{0.35\linewidth}
    \centering
    \includegraphics[width=0.35\textwidth]{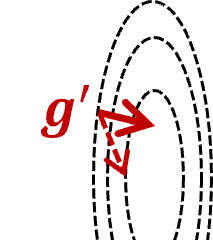}
    \caption{Landscape on a malicious client.\label{figure: malicious region}}
  \end{subfigure}
  \caption{(a) Overview of our motivating setting where benign minima (with convex hull $\mathcal{W}^*$) and the malicious minimum $\bm{w}'$ are separable. $\bm{w}$ is the parameter in the previous round, and the dashed circles in (b) and (c) are loss contours. (b) The flat landscape of a benign client (blue) along the horizontal axis reduces the horizontal gradient magnitude, allowing a malicious client (red) to easily mislead the aggregated gradient $\bar{\bm{g}}$ toward the malicious minimum $\bm{w}'$. (c) The malicious client can mimic the benign client (red dashed arrow) along the vertical dimension with less penalty due to its flat loss landscape along the vertical axis.}
  \label{figure: flatness study}
  \vspace{-0.05in}
\end{figure*}

Many recent works \citep{keskar2017on, Li2017VisualizingTL, santurkar2018does, chiang2023loss} suggest that the loss landscape of neural networks is "well-behaved" and has a flat region around the minimum (e.g., Figure 3 in \citep{Li2017VisualizingTL}). Following previous works, we discuss the difficulty of defending against federated backdoor attacks over flat loss landscapes and present concrete case studies where multiple prominent defenses can fail. Specifically, we consider a backdoor attack successful as long as the malicious clients can control the gradient direction and subsequently mislead model parameters toward the malicious minimum (Assumption \ref{assumption: separable minimum}, Figure \ref{figure: flatness study}).

% show a unique property of backdoor attacks: backdoor attacks can succeed while incurring tiny differences between malicious and benign gradients.

To begin, we formally define a flat region around a minimum $\bm{w}^*_i$ as a path-connected set (i.e., there exists at least one path that connects two points in the set) where the gradient magnitude is small. Note that a flat region may not span over the entire space but exists within a subspace, and the flatness may depend on the weight norm $\|\bm{x}\|$ \citep{Petzka2020NotesOT, Petzka2020RelativeFA}.
\begin{definition} (Flat region)
\label{definition: flat region}
    Let $\mathcal{V}$ by a subspace of the parameter space $\mathbb{R}^\mathrm{d}$, we define a $\gamma$-flat region that spans over $\mathcal{V}$ around a minimum $\bm{w}^*$ as a path-connected set $\mathcal{B}^*$ that includes $\bm{w}^*$ where the magnitude of gradient within $\mathcal{V}$ is bounded by $\gamma$: $\|\mathbb{E}_{\mathcal{D}}[\bm{g}_{\mathcal{V}}]\| \leq \gamma$.
\end{definition}

\vspace{-0.05in}
\subsection{Backdoor Attacks over a Flat Loss Landscape}
\label{section: Backdoor attacks over flat loss landscape}
\vspace{-0.05in}
The magnitude of benign gradients is small over flat loss landscapes, making it easier for the adversary to (1) mislead the aggregated gradient to the malicious minimum $\bm{w}'$ and (2) mimic benign clients to circumvent detection, e.g., by suffering a lower penalty for attack effectiveness. Figure \ref{figure: flatness study} provides some intuitive examples. In addition, the adversary can intentionally exploit the flatness property.

% \vspace{-0.05in}
\paragraph{Less dimensional perturbation requirements.}
Let $\bm{w}_{t}$ be the parameter of a global federated learning model in round $t$, where $\bm{w}_{t}$ is in regions of benign clients' loss landscapes with flatness at least $\gamma$. If a malicious client wants to guarantee that the parameter $\bm{w}_{t + 1}$ is closer to the malicious minimum $\bm{w}'$ along dimension $k$, the magnitude of its gradient $\bm{g}'$ along dimension $k$ needs to be at least $\frac{\sum_{i=1}^{N - 1}n_i}{n'} \gamma$, which decreases as the flatness increases (i.e., smaller $\gamma$). Intuitively, the more flat the loss landscapes of benign clients are, the easier it is for the malicious client to "overwrite" the aggregation result (See the horizontal axis of Figure \ref{figure: benign region} for an illustration).

Further, backdoor adversaries do not necessarily need to ``overwrite'' the aggregation result along all dimensions. Instead, backdoor attacks may perturb only a few gradient elements to minimize the overall difference between malicious and benign gradients without losing effectiveness (See the red dashed gradient in Figure \ref{figure: malicious region} for an illustration). 

\vspace{-0.05in}
\paragraph{Less penalty for mimicking benign clients.} Since the flat loss landscape is a general property of well-designed neural networks, the loss landscape of a malicious client in the unperturbed subspace can also be flat. Then, the malicious clients may partially mimic the behavior of benign clients to circumvent detection without significantly decreasing the attack success rate. Specifically, if the loss landscape of a malicious client within the unperturbed subspace is $\gamma'$-flat and the gradient magnitude of a benign client is upper bounded by $\gamma$, then it is easy to see that mimicking the benign client only decreases the effectiveness of backdoor attacks measured by the loss on backdoor samples by up to $\gamma\gamma'$, which decreases as the flatness increases (i.e., smaller $\gamma'$), via the Lagrange mean value theorem,

So far, we have seen how flat landscapes can reduce the gradient perturbation requirement of backdoor attacks and help attacks remain effective while malicious clients mimic benign clients. Even worse, an adversary may intentionally work to flatten the loss landscape, e.g., through edge-case backdoor attacks \citep{NEURIPS2020_b8ffa41d} to further increase the attack's effectiveness and circumvent defenses.

\vspace{-0.05in}
\paragraph{Edge-case attack flattens the loss landscape.} The main idea of the edge-case backdoor attack is minimizing the marginal probability of backdoor samples in the benign data distribution \citep{NEURIPS2020_b8ffa41d}. If a backdoor sample appears on both benign and malicious clients, it gets different label assignments on different types of clients. Thus, for such backdoor samples, the loss on benign clients would increase because at least one data sample in their datasets would be mispredicted. The more the loss increases, the less flat the loss landscape is, and vice versa. The edge-case backdoor attack intentionally prevents backdoor samples from appearing on benign clients and avoids the prediction error being observed to flatten loss landscapes of benign clients, as is empirically verified in Appendix \ref{appendix: experimental results}.

\begin{figure*}
  \centering
  \begin{subfigure}[b]{0.3\linewidth}
    \centering
    \includegraphics[width=.35\textwidth]{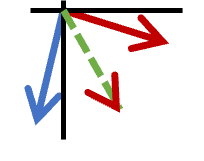}
    \caption{FLTrust failure\label{figure: FLTrust}}
  \end{subfigure}
  \hspace{1in}
  \begin{subfigure}[b]{0.3\linewidth}
    \centering
    \includegraphics[width=0.75\textwidth]{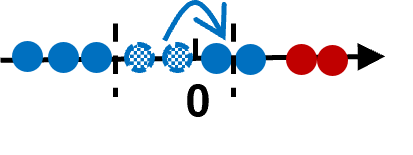}
    \caption{Median failure\label{figure: median}}
  \end{subfigure}
  \caption{Failure mode examples of existing approaches. (a) FLTrust can fail to recover the benign direction (blue) along the horizontal axis, which may subsequently converge model parameters to a malicious minimum (Figure \ref{figure: flatness study}). This is because a malicious client (red) can mimic the benign client along the vertical axis to avoid being detected as an anomaly, and misleading the aggregation result along the horizontal axis is easier due to the small benign gradient magnitude caused by flat loss landscape. (b) Median can fail to recover the benign direction (blue) even if the estimation error is small when a few benign gradients flip (blue arrow) their sign due to gradient estimation noise. Gradients with smaller magnitudes may be easier to flip for a given noise level.}
  \label{figure: failure modes}
  \vspace{-0.05in}
\end{figure*}

\vspace{-0.05in}
\subsection{Limitation of Existing Defenses over a Flat Loss Landscape}
\vspace{-0.05in}

Under the flat loss landscape setting, we discuss how existing defenses can fail to recover the correct gradient direction, including vector-wise, dimension-wise, and trigger inversion defenses. The following case study shows the failure mode of FLTrust, a vector-wise defense for federated learning systems.

% \vspace{-0.05in}
\paragraph{FLTrust.} FLTrust \citep{Cao2020FLTrustBF} uses a trusted root dataset \citep{pmlr-v97-xie19b} to generate a reference gradient vector $\bm{g}^*$. Then, FLTrust weights each reported gradient vector using its cosine similarity to the reference before normalization and aggregation. To simplify the discussion, we consider an example with one benign client whose gradient aligns with the reference gradient (i.e., cosine similarity is 1), one malicious client, a uniform sample number across clients, and two-dimensional gradient vectors. Suppose the benign client has a flat loss landscape along the first dimension (i.e., $|\bm{g}_1| \gg \gamma_0 \geq |\bm{g}_0|$). The following proposition suggests that the aggregation result always points to the direction specified by the malicious client along the first dimension (Figures \ref{figure: FLTrust}) due to the reduced perturbation requirement in a flat loss landscape setting (i.e., a small $\gamma_0$) as is discussed in Section \ref{section: Backdoor attacks over flat loss landscape}.

\begin{proposition} \label{proposition: fltrust failure}
Let $\bm{g}$ be a 2-dimensional (2-d) benign gradient, $\bm{g}'$ be a 2-d malicious gradient, and $\bm{g}^*$ be a 2-d reference gradient estimated over the trust root dataset, suppose $\bm{g}_0\bm{g}'_0 < 0$ and $\bm{g}_1\bm{g}'_1 > 0$, under the aggregation rule of FLTrust which enforces $\|\bm{g}\| = \|\bm{g}'\| =  \|\bm{g}^*\|$, if $\|\bm{g}_0\| \leq \gamma_0  \leq \|\bm{g}\| \cdot \mathrm{cos}(0.4\pi)$, there exists a malicious gradient $\bm{g}'$ such that $\|\bm{g}'_0\|\frac{\bm{g}' \cdot \bm{g}^*}{\|\bm{g}'\|\|\bm{g}^*\|} > \|\bm{g}_0\|$ and $\bm{g}'_0 (\bm{g}'_0 \frac{\bm{g}' \cdot \bm{g}^*}{\|\bm{g}'\|\|\bm{g}^*\|} + \bm{g}_0) > 0$
\end{proposition}

Meanwhile, directly applying dimension-wise defenses may not be effective either. The following example focuses on the median estimator \citep{10.1007/978-3-030-46147-8_13} from robust statistics.

% \vspace{-0.05in}
\paragraph{Median.} Robust statistics provides valid estimation results with small errors in the presence of outliers \citep{10.1214/20-AOS1961} and gradient estimator noise. However, they are not aware of the "location" of the ground-truth value. Therefore, for a median $\bm{\tilde{g}}_k$ and an estimation error $\|\bm{\tilde{g}}_k - \mathbb{E}[\bm{g}_k]\|$, if the median $\bm{\tilde{g}}_k$ has a small magnitude over a flat landscape such that $\|\bm{\tilde{g}}_k\| < \|\bm{\tilde{g}}_k - \mathbb{E}[\bm{g}_k]\|$, the adversary may mislead the aggregation result pointing to the malicious minimum. Figure \ref{figure: median} shows such a failure mode. Later, we will theoretically show that our invariant aggregator can reduce the probability of such a failure mode.

% \begin{theorem}\label{theorem: trimmed-mean error}\citep{10.1214/20-AOS1961} (Rephrased) For a uni-variate median estimator with $N$ samples, with a probability at least $1 - e^{-\frac{N}{24}}$, $\|\bm{g}_k - \mathbb{E}[\bm{g}_k]\| \leq \frac{10}{\sqrt{2}}\bm{\sigma}_k$
    
% \end{theorem}

% \begin{proposition}
%     Under Theorem \ref{theorem: trimmed-mean error}, with a probability at least ..., $\|\bm{g}_k - \mathbb{E}[\bm{g}_k]\| \leq \mathbb{E}[\bm{g}_k]$
% \end{proposition}

In addition to the vector-wise and dimension-wise defenses, trigger inversion \citep{Wang2019NeuralCI, zhang2023flip} is a more directed defense against backdoor attacks that aim to reverse engineer backdoor samples and assign them correct labels. However, we find that trigger inversion can be less effective against semantic backdoor attacks \citep{NEURIPS2020_b8ffa41d}, whose trigger does not have a fixed shape or location and is, therefore, difficult to reverse engineer precisely. The following discussion further details the limitation of trigger inversion approaches from a statistical distance perspective, showing why imprecise trigger inversion has reduced effectiveness.

% \vspace{-0.1in}

\paragraph{Trigger inversion.}
Trigger inversion approaches aim to generate a data distribution $\mathcal{D}$ that is close to the backdoor data distribution $\mathcal{D}'$ but has correct label assignments. Suppose a trigger inversion defense recovers the backdoor data distribution. In that case, any backdoor attack that misleads model predictions will be observed via evaluating models over $\mathcal{D}$, resulting in a non-flat loss landscape. However, the useful gradients may get redirected and fail to erase the backdoor as the inversed distribution $\mathcal{D}$ shifts away from the backdoor data distribution $\mathcal{D}'$, hurting the non-flatness guarantee from trigger inversion defense. The following formalizes this observation.

\begin{proposition}
    For a model with $\lambda$-Lipschitz gradient, the difference $\|\mathbb{E}_{\mathcal{D}}[\bm{g}] - \mathbb{E}_{\mathcal{D}'}[\bm{g}']\|$ between the gradient $\bm{g}$ over $\mathcal{D}$ and $\bm{g}'$ over $\mathcal{D}'$  can go up to $\lambda \mathrm{W}_1(\mathcal{D}, \mathcal{D}')$, meaning that $\gamma \geq \max\Big(0, \gamma' - \lambda \mathrm{W}_1(\mathcal{D}, \mathcal{D}')\Big)$.
\end{proposition}

% \vspace{-0.1in}
\section{Method}
% \vspace{-0.1in}
We propose an invariant aggregator to defend against backdoor attacks by avoiding the updates where the aggregated gradient benefits a few and possibly malicious clients regardless of their gradient magnitudes. Specifically, our approach considers the gradient sign as a magnitude-agnostic indicator of benefit and avoids taking any optimization steps that can not benefit sufficiently many clients. The magnitude-agnostic property maintains the robustness of our approach over flat loss landscapes where the gradient magnitude shrinks. Thus, our approach is distinct from existing approaches that consider malicious clients as anomalies or outliers \citep{Cao2020FLTrustBF, 9721118}. In what follows, we introduce the technical details of our invariant aggregator and discuss how it provably mitigates backdoor attacks by decreasing the attack success rate.

% \vspace{-0.1in}
\subsection{Invariant Aggregator}
% \vspace{-0.05in}
\paragraph{AND-Mask.}
The AND-mask \citep{parascandolo2021learning} computes a dimension-wise mask by inspecting the sign consistency of each dimension across samples. Here, we apply it to the sign consistency across clients. For dimension $k$, the sign consistency is: $|\frac{1}{N}\sum_{i=1}^{N}\mathrm{sign}(\bm{g}_{i, k})|$. If the sign consistency is below a given threshold $\tau$, the mask element $m_k$ is set to 0; otherwise, $m_k$ is set to 1. The mask along dimension $k$ is defined as:

\begin{definition} (AND-Mask)
\label{definition: and mask}
For the $k^{\mathrm{th}}$ dimension in the gradient vector, the corresponding mask $m_k$ is defined as:
% \begin{equation}
    $m_k := \frac{1}{N} \cdot \bm{1}\left[|\sum_{i=1}^{N }\mathrm{sign}(\bm{g}_{i, k})| > \tau\right]$.
% \end{equation}
% \vspace{-0.1in}
\end{definition}

Our defense then multiplies the mask $m$ with the aggregated gradient $\tilde{\bm{g}}$ element-wise, setting the inconsistent dimension to zero to avoid benefiting few malicious clients. Since the AND-mask focuses on the gradient direction, its effectiveness does not diminish due to the flatness of the landscape, which only affects the gradient magnitudes. For a consistent dimension, we call the majority gradient direction ``invariant direction''. However, if we naively average the gradient elements in consistent dimensions, malicious clients could use outliers to mislead the averaging result away from the invariant direction.

% \vspace{-0.1in}
\paragraph{Trimmed-mean.}
To complement the AND-mask and ensure that the aggregation result does follow invariant directions in consistent dimensions, our defense broadcasts the trimmed-mean estimator to each gradient dimension. The trimmed-mean estimator alleviates the outlier issue by removing the subset of the largest and smallest elements before computing the mean. The largest and smallest elements appear on the two tails of a sorted sequence. Next, we define order statistics and the trimmed mean estimator.
%, which is defined as follows.

% % \vspace{0.05in}
\begin{definition}
\label{definition: order statistics}
(Order Statistics) \citep{10.1007/978-3-030-46147-8_13} By sorting the scalar sequence $\{x_i: i \in \{1, ..., N\}, x_i \in \mathbb{R}\}$, we get $x_{1:N} \leq x_{2:N} \leq ... \leq x_{N:N}$, where $x_{i:N}$ is the $i^{\mathrm{th}}$ smallest element in $\{x_i: i \in \{1, ..., N\}\}$.
% \vspace{-0.1in}
\end{definition}
% % \vspace{0.05in}

Then, the trimmed-mean estimator removes $\alpha \times N$ elements from each tail of the sorted sequence.
% % \vspace{0.05in}

\begin{definition}
\label{definition: trimmed mean}
(Trimmed Mean Estimator) \citep{10.1007/978-3-030-46147-8_13} For $\alpha \in [0, 1]$, the $\alpha$-trimmed mean of the set of scalars ${x_{i: N} \in \{1, ..., N\}}$ is defined as:
% \begin{equation}
    $\mathrm{TrMean}(\{x_1, ..., x_{N}\}; \alpha) = \frac{1}{N - 2 \cdot \lceil \alpha\cdot N\rceil} \sum_{i=\lceil \alpha\cdot N\rceil + 1}^{N - \lceil \alpha\cdot N\rceil} x_{i:N}$,
% \end{equation}
where $\lceil . \rceil$ denotes the ceiling function.
% \vspace{-0.1in}
\end{definition}
% % \vspace{0.05in}

Algorithm \ref{algorithm: defense} outlines the steps of our server-side defense that implements invariant aggregation of gradients from the clients. The solution is composed of the AND-mask and trimmed-mean estimator. Our defense applies the two components separately based on the sign consistency of each dimension with a threshold $\tau$. 
% % \vspace{0.05in}

\begin{algorithm}[htb]
\caption{Server-side Defense}
\label{algorithm: defense}
\begin{algorithmic}[1] % Show the line index. 
\REQUIRE ~~\\ % Input
    A set of reported gradients, $\{\bm{g}_i \mid i \in \{1, ..., N\}\}$; \\
    Hyper-parameters $\tau$, $\alpha$;
\ENSURE ~~\\ % Output
    \STATE Compute $m := \frac{1}{N} \cdot\bm{1}\left[|\sum_{i=1}^{N }\mathrm{sign}(\bm{g}_i)| > \tau\right]$;
    % following Definition \ref{definition: and mask};
    \STATE Compute $\tilde{\bm{g}} := \mathrm{TrMean}(\{\bm{g}_1, ..., \bm{g}_{N}\}; \alpha)$;
    % under Definition \ref{definition: trimmed mean};
\RETURN $\bar{\bm{g}} := m \odot \tilde{\bm{g}}$;
\end{algorithmic}
\end{algorithm}

% % \vspace{0.05in}

% \vspace{-0.1in}
\subsection{Provable Mitigation}
% \vspace{-0.1in}
We demonstrate the provable backdoor mitigation of our invariant aggregator by showing that it maintains the progress of converging a model parameter $\bm{w}$ toward benign minima $\mathcal{W}^*$ and reduces the probability of moving toward the malicious minimum $\bm{w}'$(Assumption \ref{assumption: separable minimum}). Our results also suggest that (1) the invariant aggregator is more effective than baselines and (2) the flat loss landscape is less likely to break our invariant aggregation, differing from existing approaches (Section \ref{figure: failure modes}). We start with a single-dimension analysis where the benign minima $\mathcal{W}^*$ and the malicious minimum $\bm{w}'$ are on different sides of the current parameter (e.g., horizontal dimension in Figure \ref{figure: minima}).
% % \vspace{0.05in}

% \paragraph{Numerical example}
% Suppose there are ten clients in total, and two of the clients are malicious. The expected gradients of benign clients along the $k^\mathrm{th}$ dimension are positive, and those of malicious clients are negative. Under the present of gradient estimation noise (Assumption \ref{assumption: noisy gradient}), we use $p$ to denote the maximum probability of an estimated benign gradient having a negative sign. The probability $p$ may increase as the noise magnitude $\bm{\sigma}$ increases. For an arithmetic mean estimator, malicious clients can always mislead the aggregation result to negative. For a $\alpha$-trimmed-mean estimator with $\alpha \geq 2$, the aggregation result has a probability at least $$. 
% % \vspace{0.05in}

\begin{theorem} (Single-dimension)
\label{theorem: single-dimension}
    Under Assumption \ref{assumption: noisy gradient}, for a parameter $\bm{w} \notin \mathcal{W}^*$ where $(\bm{w}_k - \bm{w}^*_{i, k})(\bm{w}_k - \bm{w}'_k) \leq 0, \forall i \in \{1, ..., N - N'\}$ along the $k^{\mathrm{th}}$ dimension, 
    % suppose there are $\hat{N}$ benign clients indexed by $i$ whose gradient satisfies $\bm{g}_{i, k} (\bm{w}_k - \bm{w}^*_{i, k}) > 0$,
    let the sign-flipping probability $p_k = \max_{i \in \{1, ..., N - N'\}} \mathbb{P}[\mathbb{E}[\bm{g}_{i, k}] \bm{g}_{i, k} < 0]$ and $\bar{\bm{g}}$ be the aggregated gradient, using the invariant aggregator with $\frac{N'}{N} \leq \alpha < \frac{1}{2}$ and $\tau = 1 - 2 \alpha$, with probability at least $p_- = \sum_{i=N - \alpha N}^{N - N'} (1 - p)^i$, we have the aggregated $\bar{\bm{g}}_k$ points to the benign $\mathcal{W}^*$ and with probability at most $p_+ = \sum_{i=\frac{1 + \tau}{2}N -N' + 1}^{N - N'}p^i$ we have the aggregated $\bar{\bm{g}}_k$ points to the malicious $\bm{w}'$. In contrast, we have $p_- = \sum_{i=N - \alpha N}^{N - N'} (1 - p) ^i$ and $p_+ = \sum_{i=\alpha N - N' + 1}^{N - N'}p^i$ if we use the trimmed-mean estimator alone and have $p_- = 0$ and $p_+ = 1$ if using the arithmetic mean.
\end{theorem}
% % \vspace{0.05in}
% $p_{k, -} = \sum_{i=N - \alpha N}^{N - N'} (1 - p_k)^i$

% $p_{k, +} = \sum_{i=\frac{1 + \tau}{2}N -N' + 1}^{N - N'}p_k^i$

% $p_{k, -} = \sum_{i=N - \alpha N}^{N - N'} (1 - p_k) ^i$

% $p_{k, +} = \sum_{i=\alpha N - N' + 1}^{N - N'}p_k^i$

% $p_{k, -} = 0$

% $p_{k, +} = 1$

The high-level idea of Theorem \ref{theorem: single-dimension} is to preserve the stable progress toward benign minima (i.e., $p_-$) and cut off potential progress toward the malicious minimum (i.e., $p_+$). For example, in the case of Figure \ref{figure: median}, our invariant aggregator will not take any steps along malicious gradients due to reduced sign consistency if $\tau \geq \frac{1}{7}$. In general, Theorem \ref{theorem: single-dimension} suggests our approach is more effective than the trimmed-mean estimator in reducing the attack success rate because $\frac{1 + \tau}{2} > \alpha$. Then, we analyze the convergence guarantee of our aggregator to a neighborhood of benign minima $\mathcal{W}^*$ because our single-dimension result holds when the current parameter $\bm{w} \notin \mathcal{W}^*$ and malicious clients may pull $\bm{w}$ out of $\mathcal{W}^*$ by up to one step.
% when $\bm{w} \in \mathcal{W}^*$.

\begin{theorem} (Convergence)
\label{theorem: convergence}
    Under Assumptions \ref{assumption: bounded heterogeneity} - \ref{assumption: noisy gradient} and Theorem \ref{theorem: single-dimension}, let $\bm{w}$ be a initial parameter, suppose $|\bm{w}_k - \bm{w}^*_{i, k}| \leq c$ and $\eta_{-} \leq |\bar{\bm{g}}_{k, t}| \leq \eta^{-}$, $\forall i \in \{1, ..., N - N'\}, k \in \{1, ..., \mathrm{d}\}, t \in \{1, ...\}$, $|\bar{\bm{g}}_{k, t}| > 0$, if the number of round $T \geq \frac{c}{\eta_{-}}$, with a probability at least $\big[\sum_{i=\frac{c}{\eta_{-}}}^{T} \mathcal{F}(T, i, p_-)
    % {T \choose i} (p_-)^i(1 - p_-)^{T - i} 
    \cdot \sum_{j=\frac{i \cdot n_- - c}{n^-}}^{T-i} 
    % {T - i \choose j}(\frac{1 - p_- - p_+}{1 - p_-})^j(1 - \frac{1 - p_- - p_+}{1 - p_-})^{T - i - j}
    \mathcal{F}(T -i, j, \frac{1 - p_- - p_+}{1 - p_-})\big]^\mathrm{d}$ where $\mathcal{F}(T, i, p_-)$ denotes a binomial density function with $T$ trails, $i$ success, and probability $p_-$,
    we have $\bm{w}_T \in \{\bm{w} \mid \exists \bm{w}^* \in \mathcal{W}^*, \|\bm{w} - \bm{w}^*\| \leq \sqrt{\mathrm{d}}\eta^{-}\}$.
\end{theorem}

The proof of Theorem \ref{theorem: convergence} is straightforward: we compute the minimum number of steps that a model needs to converge the benign minima and count how many wrong steps toward the malicious minimum we can tolerate. Since the binomial cumulative density function (i.e., the sum of $\mathcal{F}$) is monotonic, we can see that the convergence probability increases with a larger $p_-$ and a smaller $p_+$. In Theorem \ref{theorem: single-dimension}, we have already shown that our approach reduces $p_+$ without hurting $p_-$. 

% \vspace{-0.1in}
\paragraph{Connection to Flatness} The loss landscape flatness still plays a role because the sign-flipping probability $p$ in Theorem \ref{theorem: single-dimension} can increase with more flatness (i.e., a smaller $\gamma$) and larger noise magnitudes $\bm{\sigma}$ (Section \ref{section: assumptions}) via Chebyschev's inequality: $\mathbb{P}(|\mathbb{E}[\bm{g}_{i, k}] - \bm{g}_{i, k}| \geq \gamma_k) \leq \frac{\sigma_{i, k}^2}{\gamma_k^2}$. An increased sign-flipping probability can subsequently increase the failure probability $p_+$ and reduce the effectiveness of our approach. However, this does not imply that our approach suffers from flat landscapes similar to existing approaches (Section \ref{section: motivating setting}) because the sign-flipping probability $p$ depends on the interaction between the flatness and the noise. In other words, obtaining a more accurate gradient estimation \citep{NIPS2013_ac1dd209} can improve our approach over flat loss landscapes.

% \vspace{-0.1in}
\paragraph{Connection to Backdoor Success.}
We further connect our convergence guarantee in Theorem \ref{theorem: convergence} to the success of backdoor attacks, which is defined via classification results. With a distribution $\mathcal{D}’$ of backdoor samples, the attack success loss $\mathcal{L}_{\mathcal{D}’}(\mathbf{w}^*)$ over a benign minimum $\mathbf{w}^* \in \mathcal{W}^*$ is expected to be high because an unperturbed benign model does not entangle backdoor triggers. With our theoretical guarantees, it is easy to lower bound the attack success loss $\mathcal{L}_{\mathcal{D}'}  (\mathbf{w}) \geq \max\Big(0, \mathcal{L} _{\mathcal{D}'}  (\mathbf{w}^*) - \gamma (\delta + \sqrt{d} \eta^{-})\Big)$ via Lagrange mean value theorem. Here, $\delta$ is the diameter of $\mathcal{W}^*$, $d$ is the dimension of the parameter space, and $\eta^{-}$ is the maximum step size defined in Theorem \ref{theorem: convergence}. 

% \vspace{-0.1in}
\paragraph{Comparison.} Our approach provides a high probability bound of mitigating backdoor attacks in Theorem \ref{theorem: convergence}. In contrast, FLTrust suffers from the failure mode in Proposition \ref{proposition: fltrust failure} and is insufficient for limiting the progress towards the malicious minimum $\bm{w}'$. In addition, the trimmed-mean defense, including the median, does not meet the same probability as our approach.

% To connect the proximity of benign minima and backdoor attacks, we first restate the separable minimum assumption (Assumption 2) in Section 3.3. We will move this assumption to the threat model. 

% Let $\mathcal{W}^*$ be a convex hull of $\{\mathbf{w}^*_i \mid i = 1, ..., N - N'\}$, we assume that malicious clients aim to let a model converge to $\mathbf{w}'$ that is not in the convex hull $\mathcal{W}^*$ of benign clients' minima. 

\begin{table*}
\centering
\caption{Accuracy of Aggregators under Continuous Edge-case Backdoor Attack. Our approach reduces the model accuracy on backdoor samples by 61.7\% on average, mitigating the backdoor attack, and achieves a comparable utility on benign samples as the standard FedAvg aggregator.}
\begin{threeparttable}
\begin{tabular}{l|@{\hspace{.5\tabcolsep}}c@{\hspace{.5\tabcolsep}}c|@{\hspace{.5\tabcolsep}}c@{\hspace{.5\tabcolsep}}c|@{\hspace{.5\tabcolsep}}c@{\hspace{.5\tabcolsep}}c|@{\hspace{.5\tabcolsep}}c@{\hspace{.5\tabcolsep}}c}
\toprule
\multirow{2}{*}{Method} &
\multicolumn{2}{c|@{\hspace{.5\tabcolsep}}}{CIFAR-10} &
\multicolumn{2}{c|@{\hspace{.5\tabcolsep}}}{Twitter} &
\multicolumn{2}{c|@{\hspace{.5\tabcolsep}}}{Phishing} 
\\
& Acc & ASR & Acc & ASR & Acc & ASR \\
\midrule
FedAvg & .679 {\tiny $\pm$ .001} & .717 {\tiny $\pm$ .001} & .722 {\tiny $\pm$ .001} & .440 {\tiny $\pm$ .001} & .999 {\tiny $\pm$ .001} & .999 {\tiny $\pm$ .001} \\
Krum \citep{Blanchard2017MachineLW} & .140 {\tiny $\pm$ .001} & .275 {\tiny $\pm$ .012} & .579 {\tiny $\pm$ .001} & .766 {\tiny $\pm$ .002} & .999 {\tiny $\pm$ .001} & .999 {\tiny $\pm$ .001} \\
Multi-Krum & .541{\tiny $\pm$ .002} & .923 {\tiny $\pm$ .021} & .727 {\tiny $\pm$ .001} & .656 {\tiny $\pm$ .008} & .999 {\tiny $\pm$ .001} & .999 {\tiny $\pm$ .001} \\
Multi-Krum$_{C}$ & .681 {\tiny $\pm$ .002} & .821 {\tiny $\pm$ .001} & .594 {\tiny $\pm$ .002} & .701 {\tiny $\pm$ .001} & .999 {\tiny $\pm$ .001} & .333 {\tiny $\pm$ .333} \\
Trimmed-Mean \citep{10.1007/978-3-030-46147-8_13} & .687 {\tiny $\pm$ .001} & .512 {\tiny $\pm$ .002} & .728 {\tiny $\pm$ .001} & .640 {\tiny $\pm$ .016} & .999 {\tiny $\pm$ .001} & .999 {\tiny $\pm$ .001} \\
Krum Trimmed-Mean & .682 {\tiny $\pm$ .001} & .607 {\tiny $\pm$ .002} & .727 {\tiny $\pm$ .001} & .641 {\tiny $\pm$ .001} & .999 {\tiny $\pm$ .001} & .999 {\tiny $\pm$ .001} \\
Sign-SGD \citep{Bernstein2019signSGDWM} & .301 {\tiny $\pm$ .005} & \textbf{.000 {\tiny $\pm$ .001}} & .610 {\tiny $\pm$ .003} & .751 {\tiny $\pm$ .076} & .999 {\tiny $\pm$ .000} & .667 {\tiny $\pm$ .333} \\
Weak-DP \citep{Sun2019CanYR} & .454 {\tiny $\pm$ .003} & .828 {\tiny $\pm$ .003} & .667 {\tiny $\pm$ .001} & .374 {\tiny $\pm$ .002} & .999 {\tiny $\pm$ .001} & .999 {\tiny $\pm$ .001} \\
% Freezing Layers & .415 {\tiny $\pm$ .001} & .572 {\tiny $\pm$ .002} & N\textbackslash A & N\textbackslash A & N\textbackslash A & N\textbackslash A \\
RLR \citep{ozdayi2021defending} & .659 {\tiny $\pm$ .001} & .450 {\tiny $\pm$ .001} & .639 {\tiny $\pm$ .002} & .447 {\tiny $\pm$ .001} & .999 {\tiny $\pm$ .001} & .999 {\tiny $\pm$ .001} \\
% FoolsGold & .667 {\tiny $\pm$ .001} & .109 {\tiny $\pm$ .001} & .726 {\tiny $\pm$ .002} & .357 {\tiny $\pm$ .001} & .999 {\tiny $\pm$ .001} & .999 {\tiny $\pm$ .001} \\
RFA \citep{9721118} & .685 {\tiny $\pm$ .001} & .853 {\tiny $\pm$ .002} & .718 {\tiny $\pm$ .001} & .704 {\tiny $\pm$ .002} & .999 {\tiny $\pm$ .001} & .999 {\tiny $\pm$ .001} \\
SparseFed \citep{pmlr-v151-panda22a} & .662 {\tiny $\pm$ .001} & .984 {\tiny $\pm$ .001} & .667 {\tiny $\pm$ .001} & .608 {\tiny $\pm$ .002} & .999 {\tiny $\pm$ .001} & .999 {\tiny $\pm$ .001} \\
FLTrust \citep{Cao2020FLTrustBF} & .671 {\tiny $\pm$ .001} & .574 {\tiny $\pm$ .002} & .691 {\tiny $\pm$ .001} & .473 {\tiny $\pm$ .002} & .999 {\tiny $\pm$ .001} & .333 {\tiny $\pm$ .001} \\
FLIP \citep{zhang2023flip} & .667 {\tiny $\pm$ .002} & .250 {\tiny $\pm$ .001} & N\textbackslash A & N\textbackslash A & .999 {\tiny $\pm$ .001} & \textbf{.000 {\tiny $\pm$ .001}} \\
\hline
No Attack & .718 {\tiny $\pm$ .001} & .000 {\tiny $\pm$ .001} & .731 {\tiny $\pm$ .001} & .095 {\tiny $\pm$ .001} & .999 {\tiny $\pm$ .001} & .000 {\tiny $\pm$ .001} \\
\hline
\textbf{Ours} & .677 {\tiny $\pm$ .001} & \textbf{.001 {\tiny $\pm$ .001}} & .687 {\tiny $\pm$ .001} & \textbf{.296 {\tiny $\pm$ .003}} & .999 {\tiny $\pm$ .001} & \textbf{.000 {\tiny $\pm$ .001}} \\
\bottomrule
\multicolumn{7}{c}{\footnotesize Note: The numbers are average accuracy over three runs. Variance is rounded up.}
\vspace{-0.2in}
\end{tabular}
\end{threeparttable}
\label{table: empirical results}
\end{table*}

% \vspace{-0.1in}
\section{Experiments}
% \vspace{-0.1in}

% Our experiment starts with synthetic data to verify our theoretical results in Section \ref{section: analysis}. Then, 
We extensively evaluate our defense on three realistic tasks with diverse data modalities and against multiple state-of-the-art backdoor attacks \citep{NEURIPS2020_b8ffa41d, Xie2020DBA, pmlr-v151-panda22a} with pixel, semantic visual, text, and value backdoor triggers.

% \vspace{-0.1in}

\paragraph{Additional Results.} Due to the limited space, (1) the loss landscape visualization, (2) an ablation study, (3) an evaluation of the hyper-parameter sensitivity, (4) empirical verifications of the separability assumption (Assumption \ref{assumption: separable minimum}), and (5) empirical evaluations under multiple state-of-the-art backdoor attack strategies \citep{NEURIPS2020_b8ffa41d, Xie2020DBA, pmlr-v151-panda22a} are in Appendix \ref{appendix: experimental results}.
% \vspace{-0.05in}

% \vspace{-0.05in}
\subsection{Experimental Setup}
\label{section: setup}
% \vspace{-0.05in}
We briefly summarize our setup and report more details in Appendix \ref{appendix: experimental setup}. 

% \vspace{-0.05in}
\textbf{Metrics.} We employ two metrics: the main task accuracy (Acc) estimated on the benign samples and the backdoor attack success rate (ASR) measured by the accuracy over backdoor samples.

% \vspace{-0.05in}
\textbf{Datasets.}
The visual data of the object detection task and text data of the sentiment analysis task are from CIFAR-10 \citep{Krizhevsky2009LearningML, pmlr-v54-mcmahan17a} and Twitter \citep{Caldas2018LEAFAB}, respectively. Each phishing email data sample has 45 standardized numerical features and binary labels (Appendix \ref{appendix: phishing dataset})

% \vspace{-0.05in}
\textbf{Federated Learning Setup.}
We consider horizontal federated learning \citep{Kairouz2021AdvancesAO} where the clients share the same feature and label spaces but non-i.i.d. data distributions. The number of clients is 100 for the three tasks. The server samples 20 clients at each round on the CIFAR-10 and phishing email experiments. Due to hardware memory limitations, we reduce the sampled client number to 15 with Twitter.

% \paragraph{CIFAR-10}
% \paragraph{Twitter}
% Each data sample is a sentence. The label is binary. y=0 means negative sentiment. We follow the CMU LEAF project on setting up the federated Sent-140 dataset. In our setting, each user has one sentence, and each client has 500 users. There are 100 clients, so the total number of samples is 50,000. The Sent-140 dataset has 240,000 samples. The test set comes from a different set of 5000 users (due to an MPI issue. If the testset is too large, the MPI buffer can overflow). All the samples are converted to embedding using GloVe-6B with 300 dimensions.

% The backdoor attack tries to create a correlation between "yorgos lanthimos" and the negative label. (See examples.)

% We use a sequence model with a two-layer 200 dimension LSTM and a 200 dimension linear layer. The optimizer is Adam (SGD does not work).

% \paragraph{Phishing}
% Constructing backdoor samples can be tricky.
% Dataset: CIFAR-10, Twitter, Phishing.
% \vspace{-0.05in}
\textbf{Backdoor Attack Setup.} We adopt edge-case \citep{NEURIPS2020_b8ffa41d}, distributed \citep{Xie2020DBA}, and collude \citep{pmlr-v151-panda22a} backdoor strategies. The pixel, semantic visual, and text backdoor trigger follow the previous work \citep{NEURIPS2020_b8ffa41d, Xie2020DBA}. For the tabular data, we select the 38$^\mathrm{th}$ feature (reputation), whose value is 0 on most of the data samples. Then, we let the adversary manipulate the 38$^\mathrm{th}$ feature to 0.2, which has a low probability density on phishing emails, and flip the label to non-phishing.

% \vspace{-0.05in}
\textbf{Adversary Setup.}
We employ strong adversaries that \emph{continuously} participate in the training \citep{9833647, zhang2023flip} and can control 20\% clients on the CIFAR-10 and phishing email experiments and 10\% clients on the Twitter experiment.

% \vspace{-0.1in}
\subsection{Result and Comparison to Prior Works}
% \vspace{-0.1in}
\label{section: empirical results} 
% \textbf{Our results.}
Table \ref{table: empirical results} summarizes the performance of each defense on three tasks against edge-case backdoor attacks that can intentionally flatten loss landscapes (Section \ref{section: motivating setting}). Our approach decreases the ASR by 61.6\% on average, outperforming all competitors. The edge-case backdoor attack on the text sentiment analysis task (Twitter) is more difficult to defend, and our approach reduces the ASR by 41.7\%. We hypothesize that the text sentiment analysis task has few invariances among clients. For example, the shape features \citep{Sun2021CanSS} in object classification tasks can be invariant across objects. In contrast, the sentiment largely depends on the entire sentence instead of a few symbols or features. We further discuss the limitations of prior defenses and evaluate our defense against various attacks in Appendix \ref{appendix: experimental results}.

\section{Conclusion and Future Work}
% \vspace{-0.1in}
This paper shows a defense against backdoor attacks by focusing on the invariant directions in the model optimization trajectory. Enforcing the model to follow the invariant direction requires AND-mask to compute the sign-consistency of each gradient dimension, which estimates how invariant a dimension-wise direction can be, and use the trimmed-mean estimator to guarantee the model follows the invariant direction within each dimension. Both theoretical and empirical results demonstrate the combination of AND-mask and the trimmed-mean estimator is necessary and effective. 

In addition, our work reveals a potential downside of flat loss landscapes, which are considered beneficial in prior work \citep{keskar2017on, foret2021sharpnessaware}. Specifically, few existing works consider the invariance of the flatness property, i.e., whether every data sample sees a flat landscape. Therefore, further combining distributional robust optimization \citep{10.1214/20-AOS2004, Sagawa*2020Distributionally} and sharpness-aware minimization \citep{foret2021sharpnessaware} to discover invariant flat minima can be interesting for future work.

% \sk{Add something about flat minima and potential future work}

\subsubsection*{Acknowledgements}
XW is partially supported by Microsoft Research and a research grant from C3.ai digital transformation institute. This work of SK is partially supported by NSF III 2046795, IIS 1909577, CCF 1934986, NIH 1R01MH116226-01A, NIFA award 2020-67021-32799, the Alfred P. Sloan Foundation, and Google Inc.

% All acknowledgments go at the end of the paper, including thanks to reviewers who gave useful comments, to colleagues who contributed to the ideas, and to funding agencies and corporate sponsors that provided financial support. 
% To preserve the anonymity, please include acknowledgments \emph{only} in the camera-ready papers.

% \clearpage
\bibliographystyle{unsrtnat}
\bibliography{main}

%%%%%%%%%%%%%%%%%%%%%%%%%%%%%%%%%%%%%%%%%%%%%%%%%%%%%%%%%%%%
\clearpage
\section*{Checklist}

% %%% BEGIN INSTRUCTIONS %%%
% The checklist follows the references. For each question, choose your answer from the three possible options: Yes, No, Not Applicable. You are encouraged to include a justification to your answer, either by referencing the appropriate section of your paper or providing a brief inline description (1-2 sentences). 
% Please do not modify the questions. Note that the Checklist section does not count towards the page limit. Not including the checklist in the first submission won't result in desk rejection, although in such case we will ask you to upload it during the author response period and include it in camera ready (if accepted).

% \textbf{In your paper, please delete this instructions block and only keep the Checklist section heading above along with the questions/answers below.}
% %%% END INSTRUCTIONS %%%

 \begin{enumerate}

 \item For all models and algorithms presented, check if you include:
 \begin{enumerate}
   \item A clear description of the mathematical setting, assumptions, algorithm, and/or model. [Yes]
   \item An analysis of the properties and complexity (time, space, sample size) of any algorithm. [Yes]
   \item (Optional) Anonymized source code, with specification of all dependencies, including external libraries. [Yes]
 \end{enumerate}

 \item For any theoretical claim, check if you include:
 \begin{enumerate}
   \item Statements of the full set of assumptions of all theoretical results. [Yes]
   \item Complete proofs of all theoretical results. [Yes]
   \item Clear explanations of any assumptions. [Yes]     
 \end{enumerate}

 \item For all figures and tables that present empirical results, check if you include:
 \begin{enumerate}
   \item The code, data, and instructions needed to reproduce the main experimental results (either in the supplemental material or as a URL). [Yes]
   \item All the training details (e.g., data splits, hyperparameters, how they were chosen). [Yes]
         \item A clear definition of the specific measure or statistics and error bars (e.g., with respect to the random seed after running experiments multiple times). [Yes]
         \item A description of the computing infrastructure used. (e.g., type of GPUs, internal cluster, or cloud provider). [Yes]
 \end{enumerate}

 \item If you are using existing assets (e.g., code, data, models) or curating/releasing new assets, check if you include:
 \begin{enumerate}
   \item Citations of the creator If your work uses existing assets. [Yes]
   \item The license information of the assets, if applicable. [Not Applicable]
   \item New assets either in the supplemental material or as a URL, if applicable. [Not Applicable]
   \item Information about consent from data providers/curators. [Not Applicable]
   \item Discussion of sensible content if applicable, e.g., personally identifiable information or offensive content. [Not Applicable]
 \end{enumerate}

 \item If you used crowdsourcing or conducted research with human subjects, check if you include:
 \begin{enumerate}
   \item The full text of instructions given to participants and screenshots. [Not Applicable]
   \item Descriptions of potential participant risks, with links to Institutional Review Board (IRB) approvals if applicable. [Not Applicable]
   \item The estimated hourly wage paid to participants and the total amount spent on participant compensation. [Not Applicable]
 \end{enumerate}

 \end{enumerate}

\clearpage
\onecolumn
\appendix
% \aistatstitle{Appendix}
\section{Table of Notations}
\begin{table}[htb]

\caption{Table of Notations}
\begin{center}
\begin{tabular}{ll}
\multicolumn{1}{c}{\bf Symbol}  &\multicolumn{1}{c}{\bf Description}
\\ \hline \\
$\bm{x}, y$ &A pair of data sample and label \\
$\mathrm{sign}(\cdot)$ &An element-wise sign operator \\
$\bm{w}$ &Parameters of the global federated learning model \\
$\bm{w}_i$ &Parameters of the $i^\text{th}$ local federated learning model \\
% $\langle  \,,  \rangle$ &An inner product of two vectors \\
% $d$, $d_r$, $d_s$ &The dimension of $\bm{x}$, $\bm{x}_r$, $\bm{x}_s$, respectively \\
$\bm{g}$ &Client update (i.e., pseudo-gradient or gradient) \\
$N$ &The number of clients \\
$N'$ &The number of malicious clients \\
$T$ &The number of training rounds \\
$\mathrm{W}_1(\cdot, \cdot)$ &Wasserstain-1 distance between two distributions \\
% $\mathrm{diag}()$ &A diagonal operator
\end{tabular}
\end{center}
\label{table: notations}
\end{table}

\section{Proofs}
\label{appendix: proofs}
\setcounter{theorem}{4}

\begin{proposition}
Let $\bm{g}$ be a 2-dimensional (2-d) benign gradient, $\bm{g}'$ be a 2-d malicious gradient, and $\bm{g}^*$ be a 2-d reference gradient estimated over the trust root dataset, suppose $\bm{g}_0\bm{g}'_0 < 0$ and $\bm{g}_1\bm{g}'_1 > 0$, under the aggregation rule of FLTrust which enforces $\|\bm{g}\| = \|\bm{g}'\| =  \|\bm{g}^*\|$, if $|\bm{g}_0| \leq \|\bm{g}\| \cdot \mathrm{cos}(0.4\pi)$, there exists a malicious gradient $\bm{g}'$ such that $|\bm{g}'_0|\frac{\bm{g}' \cdot \bm{g}^*}{\|\bm{g}'\|\|\bm{g}^*\|} > |\bm{g}_0|$ and $\bm{g}'_0 (\bm{g}'_0 \frac{\bm{g}' \cdot \bm{g}^*}{\|\bm{g}'\|\|\bm{g}^*\|} + \bm{g}_0) > 0$
\end{proposition}

\begin{figure}[h!]
    \centering
    \includegraphics[width=0.33\linewidth]{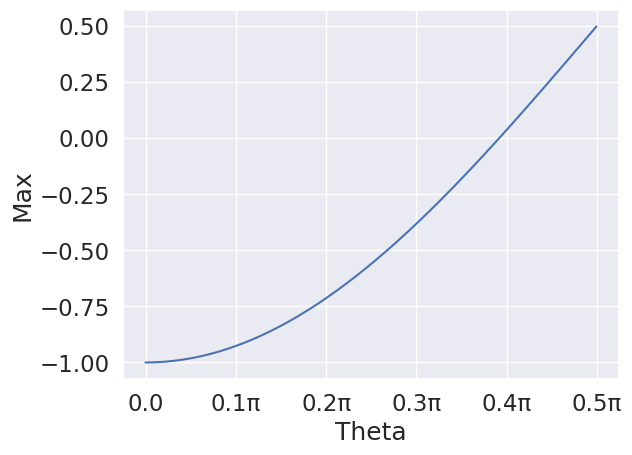}
    \caption{Numerical simulation for Equation \ref{equation: fltrust}.}
    \label{figure: numerical simulation}
\end{figure}

\begin{proof}
    Let $\theta \in [0, \frac{\pi}{2}]$ be the angle between $\bm{g}$ and $[\bm{g}_0, 0]$ and $\theta' \in [0, \frac{\pi}{2}]$ be the angle between $\bm{g}'$ and $[\bm{g}'_0, 0]$, with $\|\bm{g}\| = \|\bm{g}'\| =  \|\bm{g}^*\|$, we have $|\bm{g}'_0|\frac{\bm{g}' \cdot \bm{g}^*}{\|\bm{g}'\|\|\bm{g}^*\|} = \|\bm{g}^*\|\mathrm{cos}(\theta')\mathrm{cos}(\pi - \theta - \theta')$ and $|\bm{g}_0| = \|\bm{g}^*\|\mathrm{cos}\theta$. Then, $|\bm{g}'_0|\frac{\bm{g}' \cdot \bm{g}^*}{\|\bm{g}'\|\|\bm{g}^*\|} > |\bm{g}_0|$ is equivalent to 
    \begin{equation}
    \label{equation: fltrust}
        \mathrm{cos}(\theta')\mathrm{cos}(\pi - \theta - \theta') - \mathrm{cos}\theta > 0.
    \end{equation}
    We are interested in figuring out, for a given benign gradient with a fixed $\theta$, whether there exists a malicious gradient with $\theta'$ such that Equation \ref{equation: fltrust} holds. Since an analytical solution for Equation \ref{equation: fltrust} may not be tractable, we resort to the numerical simulation. Figure \ref{figure: numerical simulation} shows the plot of $\max_{\theta' \in [0, \frac{\pi}{2}]} \mathrm{cos}(\theta')\mathrm{cos}(\pi - \theta - \theta') - \mathrm{cos}\theta$ for each given $\theta$ in $[0, \frac{\pi}{2}]$. It is easy to see that as long as $\theta \geq 0.4\pi$, there exists a $\theta' \in [0, \frac{\pi}{2}]$ such that Equation \ref{equation: fltrust} holds.
\end{proof}

\begin{proposition}
    For a model with $\lambda$-Lipschitz gradient, the difference $\|\mathbb{E}_{\mathcal{D}}[\bm{g}] - \mathbb{E}_{\mathcal{D}'}[\bm{g}']\|$ between the gradient $\bm{g}$ over $\mathcal{D}$ and $\bm{g}'$ over $\mathcal{D}'$  can go up to $\lambda \mathrm{W}_1(\mathcal{D}, \mathcal{D}')$.
\end{proposition}

\begin{proof}
    Let $\bm{g}(\bm{x})$ denotes a gradient at $\bm{x}$ and $p(\bm{x}, \bm{x}')$ be the joint probability of $\bm{x}$ and $\bm{x}'$, we have
    \begin{equation}
    \begin{split}
        \|\mathbb{E}_{\mathcal{D}}[\bm{g}] - \mathbb{E}_{\mathcal{D}'}[\bm{g}']\| &= \|\mathbb{E}_{\mathcal{D}}[\bm{g}(\bm{x})] - \mathbb{E}_{\mathcal{D}'}[\bm{g}'(\bm{x}')]\| \\
        &\leq \lambda \int \|\bm{x} - \bm{x'}\| p(\bm{x}, \bm{x}') \\
        &\leq \lambda \mathrm{W}_1(\mathcal{D}, \mathcal{D}')
    \end{split}
    \end{equation}
\end{proof}

\setcounter{theorem}{9}

\begin{theorem} (Single-dimension)
    Under Assumption \ref{assumption: noisy gradient}, for a parameter $\bm{w} \notin \mathcal{W}^*$ where $(\bm{w}_k - \bm{w}^*_{i, k})(\bm{w}_k - \bm{w}'_k) \leq 0, \forall i \in \{1, ..., N - N'\}$ along the $k^{\mathrm{th}}$ dimension, 
    % suppose there are $\hat{N}$ benign clients indexed by $i$ whose gradient satisfies $\bm{g}_{i, k} (\bm{w}_k - \bm{w}^*_{i, k}) > 0$,
    let the sign-flipping probability $p = \max_{i \in \{1, ..., N - N'\}} \mathbb{P}[\mathbb{E}[\bm{g}_{i, k}] \bm{g}_{i, k} < 0]$ and $\bar{\bm{g}}$ be the aggregated gradient, using the invariant aggregator with $\frac{N'}{N} \leq \alpha < \frac{1}{2}$ and $\tau = 1 - 2 \alpha$, with probability at least $p_- = \sum_{i=N - \alpha N}^{N - N'} (1 - p)^i$, we have the aggregated $\bar{\bm{g}}_k$ points to the benign $\mathcal{W}^*$ and with probability at most $p_+ = \sum_{i=\frac{1 + \tau}{2}N -N' + 1}^{N - N'}p^i$ we have the aggregated $\bar{\bm{g}}_k$ points to the malicious $\bm{w}'$. In contrast, we have $p_- = \sum_{i=N - \alpha N}^{N - N'} (1 - p) ^i$ and $p_+ = \sum_{i=\alpha N - N' + 1}^{N - N'}p^i$ if we use the trimmed-mean estimator alone and have $p_- = 0$ and $p_+ = 1$ if using the arithmetic mean estimator with or without the AND-mask.
\end{theorem}

\begin{proof}
    Under the given setup, a sufficient condition for the trimmed-mean pointing toward the benign direction is that the remaining elements that are not trimmed point toward the benign direction. Since the trim threshold $\alpha$ is greater or equal to the corruption level $\frac{N'}{N}$, guaranteeing at least $N - \alpha N$ benign elements point to the benign direction is sufficient for guaranteeing the direction of the remaining elements after trimming. In addition, setting the masking threshold $\tau = 1 - 2\alpha$ guarantees that a dimension with at least $N - \alpha N$ benign elements pointing to the benign direction will not be masked out. With the definition of sign flipping probability $p$, we have $p_- = \sum_{i=N - \alpha N}^{N - N'} (1 - p)^i$ for our invariant aggregator and the trimmed-mean estimator. In contrast, for the arithmetic mean estimator, malicious clients can always use malicious gradient elements that are greater than the sum of benign gradient elements to control the aggregation result. Therefore, we have $p_- = 0$ for the arithmetic mean estimator.

    Similarly, a dimension-wise aggregation result may align with the malicious gradient if there exists a remaining element that points to the malicious direction after trimming. Therefore, for the trimmed mean estimator, its failure rate is at most $p_+ = \sum_{i=\alpha N - N' + 1}^{N - N'}p^i$. 
    
    However, with AND-mask, only one remaining element pointing to the malicious direction after trimming may not be sufficient for reaching a high sign consistency to pass the mask. 
    For an invariant aggregator with a masking threshold $\tau$, there needs at least $\frac{1 + \tau}{2}N -N' + 1$ benign gradient elements flipping their directions and align with the malicious gradient elements, resulting in a smaller failure rate upper bound $p_+ = \sum_{i=\frac{1 + \tau}{2}N -N' + 1}^{N - N'}p^i$. 

    This is because if there is less than $\frac{1 + \tau}{2}N -N' + 1$ benign update elements flipping their directions and aligning with the malicious update elements, such a dimension (1) will be set to 0 due to low sign consistency or (2) passes the AND-mask and the trimmed-mean estimator recovers the benign gradient direction. In contrast, without masking, there only needs $\alpha N – N’ + 1$ flipped benign update to cause a failure mode of progressing toward malicious minimum (Figure \ref{figure: median}). Note that $\alpha < \frac{1 + \tau}{2}$. Without trimming, a single malicious update may manipulate the arithmetic mean to an arbitrary value in a consistent dimension.
\end{proof}

\begin{theorem} (Convergence)
    Under Assumptions \ref{assumption: bounded heterogeneity} - \ref{assumption: noisy gradient} and Theorem \ref{theorem: single-dimension}, let $\bm{w}$ be a initial parameter, suppose $|\bm{w}_k - \bm{w}^*_{i, k}| \leq c$ and $\eta_{-} \leq |\bar{\bm{g}}_{k, t}| \leq \eta^{-}$, $\forall i \in \{1, ..., N - N'\}, k \in \{1, ..., \mathrm{d}\}, t \in \{1, ...\}$, $|\bar{\bm{g}}_{k, t}| > 0$, if the number of round $T \geq \frac{c}{\eta_{-}}$, with a probability at least $\big[\sum_{i=\frac{c}{\eta_{-}}}^{T} \mathcal{F}(T, i, p_-)
    % {T \choose i} (p_-)^i(1 - p_-)^{T - i} 
    \cdot \sum_{j=\frac{i \cdot n_- - c}{n^-}}^{T-i} 
    % {T - i \choose j}(\frac{1 - p_- - p_+}{1 - p_-})^j(1 - \frac{1 - p_- - p_+}{1 - p_-})^{T - i - j}
    \mathcal{F}(T -i, j, \frac{1 - p_- - p_+}{1 - p_-})\big]^\mathrm{d}$ where $\mathcal{F}(T, i, p_-)$ denotes a binomial density function with $T$ trails, $i$ success, and probability $p_-$,
    we have $\bm{w}_T \in \{\bm{w} \mid \exists \bm{w}^* \in \mathcal{W}^*, \|\bm{w} - \bm{w}^*\| \leq \sqrt{\mathrm{d}}\eta^{-}\}$.
\end{theorem}

\begin{proof}
    Under the given conditions, an optimization strategy needs at least $\frac{c}{\eta_-}$ steps to converge an initial parameter $\bm{w}$ to a minimum $\bm{w}^*_i$ for any given $i \in \{1, ..., N - N'\}$. Therefore, we need to number of rounds $T \geq \frac{c}{\eta_-}$. Then, using Theorem \ref{theorem: single-dimension}, for a given $T$, within a certain dimension, we have a probability at least $\sum_{i=\frac{c}{\eta_{-}}}^{T}\mathcal{F}(T, i, p_-)$ the parameter $\bm{w}_T$ can possibly reach the benign minimum, where $\mathcal{F}(T, i, p_-)$ denotes a binomial density function with $T$ trails, $i$ success, and probability $p_-$. Here, we are not absolutely certain about the convergence because malicious clients may also mislead the parameter and move away from the benign minima.

    To further incorporate malicious clients into consideration, we analyze how many steps in an optimization trajectory can be misled by malicious clients without hurting the convergence to benign minima. Suppose the parameter steps toward benign minima for $i$ steps. We need at least $\frac{i \cdot n_- - c}{n^-}$ steps among the remaining $T - i$ steps to not direct to the malicious minimum, whose probability is at least $\sum_{j=\frac{i \cdot n_- - c}{n^-}}^{T-i} \mathcal{F}(T -i, j, \frac{1 - p_- - p_+}{1 - p_-})$.

    In addition, if malicious clients mislead the aggregation result in the last round, we will not be able to converge the parameter to the benign minima. Therefore, we relax the convergence to the benign minima to the convergence to a neighborhood around the benign minima, whose radius is bounded by the gradient norm $\sqrt{\mathrm{d}}\eta^{-}$.

    Combining $\sum_{i=\frac{c}{\eta_{-}}}^{T}\mathcal{F}(T, i, p_-)$ and $\sum_{j=\frac{i \cdot n_- - c}{n^-}}^{T-i} \mathcal{F}(T -i, j, \frac{1 - p_- - p_+}{1 - p_-})$ and taking the product across all dimensions complete the proof.
\end{proof}

\section{Experimental Setup}
\label{appendix: experimental setup}

\subsection{Backdoor Attacks}
The edge-case backdoor attack \citep{NEURIPS2020_b8ffa41d}, distributed backdoor attack \citep{Xie2020DBA}, and colluding attack \citep{pmlr-v151-panda22a} follow previous works. For the adaptive adversary, we let it intentionally scale up the gradient elements in dimensions that are not masked out such that the gradient norms of the masked malicious gradients remain the same.

\subsection{Model Architectures}
We use a Resnet-18 \citep{He2016DeepRL} on the CIFAR-10 dataset. The GloVe-6B \citep{pennington-etal-2014-glove} provides word embedding for the Twitter dataset, and a two-layer LSTM model with 200 hidden units further uses the embedding for sentiment prediction. A three-layer 128-256-2 fully connected network is employed to detect phishing emails.

\subsection{Hyper-parameters}
Table \ref{table: hyper-params} lists the hyper-parameters. We start with $\tau = \frac{N'}{N}$ and increase $\tau$ if there needs more robustness.

\begin{table*}[h!]
  \caption{Hyper-parameters}
  \centering
  \begin{tabular}{lccc}
    \toprule
    Hyper-parameters & CIFAR-10 & Twitter & Phishing \\
    \midrule
    Optimizer           & SGD & Adam & SGD \\
    Learning Rate       & 0.01 & 0.0001 & 0.1 \\
    Batch Size          & 64 & 64 & 64 \\
    % Number of Backdoor Samples per Batch & 64 & 64 & 64 \\
    Local Epoch         & 1 & 0.1 & 1 \\
    Communication Round & 300 & 300 & 20 \\
    Number of Clients & 100 & 100 & 100 \\
    Number of Malicious Clients & 20 & 10 & 20 \\
    Number of Clients per Round & 20 & 15 & 20 \\
    $\tau$ & 0.2 & 0.6 & 0.6 \\
    $\alpha$ & 0.25 & 0.25 & 0.25 \\
    \bottomrule
    \label{table: hyper-params}
  \end{tabular}
\end{table*}

\subsection{Phishing Dataset}\label{appendix: phishing dataset}
Each phishing email data sample has 45 standardized numerical features of the sender that represent the sender's reputation scores. A large reputation score may indicate a phishing email. The reputation scores come from peer-reviewers in a reputation system \citep{Jsang2007ASO}. The adversary may use malicious clients to manipulate the reputation.

\section{Experimental Results}
\label{appendix: experimental results}
\subsection{Loss Landscape Visualization}

\begin{figure*}[h!]
  \centering
  \begin{subfigure}[c]{0.2125\linewidth}
    \centering\includegraphics[width=\textwidth]{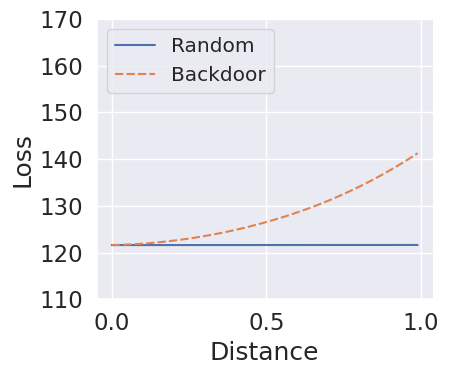}
    \caption{Edge-case 0\%}
  \end{subfigure}
  \hspace{0.2in}
  \begin{subfigure}[c]{0.2125\linewidth}
    \centering\includegraphics[width=\textwidth]{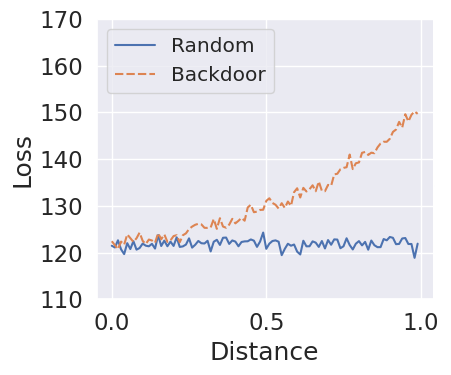}
    \caption{Edge-case 15\%}
  \end{subfigure}
  \hspace{0.2in}
  \begin{subfigure}[c]{0.2125\linewidth}
    \centering\includegraphics[width=\textwidth]{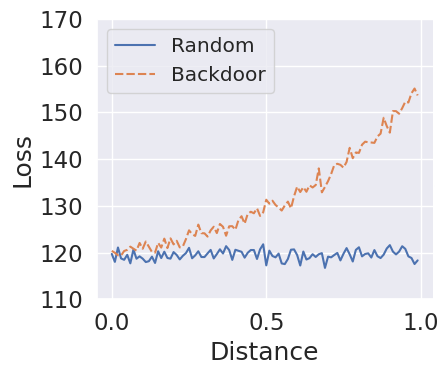}
    \caption{Edge-case 30\%}
  \end{subfigure}
  \hspace{0.2in}
  \begin{subfigure}[c]{0.2125\linewidth}
    \centering\includegraphics[width=\textwidth]{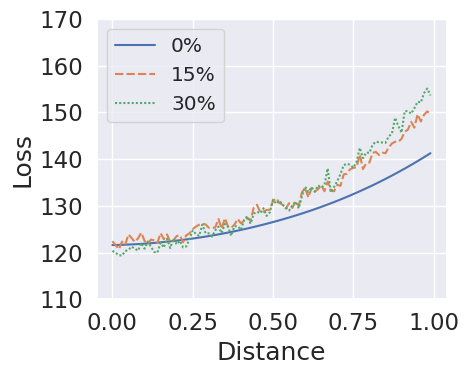}
    \caption{Comparison}
  \end{subfigure}
  \caption{Loss landscapes along the direction of malicious and random gradients. Distance is defined using the gradient norm. Distance 0.0 means the parameter stays at a minimum.}
  \label{figure: loss landscape}
\end{figure*}

Due to the high dimensionality of model parameters, we shall focus on the loss landscape along the gradient directions. Specifically, we consider backdoor gradient directions and random gradient directions as comparisons. Figure \ref{figure: loss landscape} shows two loss landscapes with different edge-case backdoor attack configurations, whose backdoor samples have probabilities of 0\%, 15\%, and 30\% to appear on benign data distributions, respectively. Here, the edge-case 0\% attack has a most flat loss landscape, complementing our analysis in Section \ref{section: Backdoor attacks over flat loss landscape}. We kindly refer to related studies \citep{zhou2018sgd, pmlr-v80-kleinberg18a} on epoch-wise convexity for readers that wonder why SGD can escape from the bad minima in the edge-case 15\% and 30\% loss landscapes.

\subsection{Mimicking Benign Clients}
\begin{figure}[h!]
  \centering
  \begin{subfigure}[c]{0.33\linewidth}
    \centering\includegraphics[width=\textwidth]{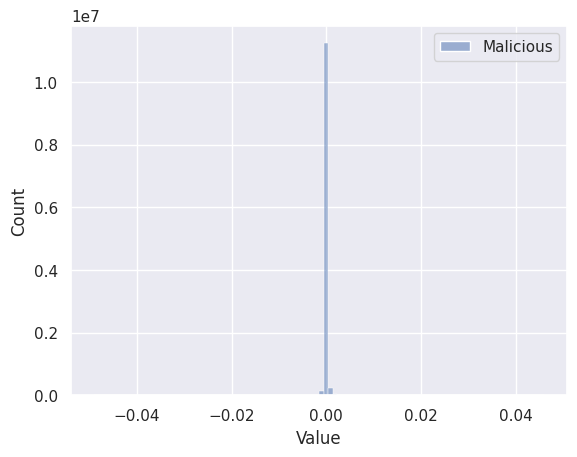}
    \caption{Gradient element value histogram}
  \end{subfigure}
  % \hspace{1.0in}
  \begin{subfigure}[c]{0.33\linewidth}
    \centering\includegraphics[width=\textwidth]{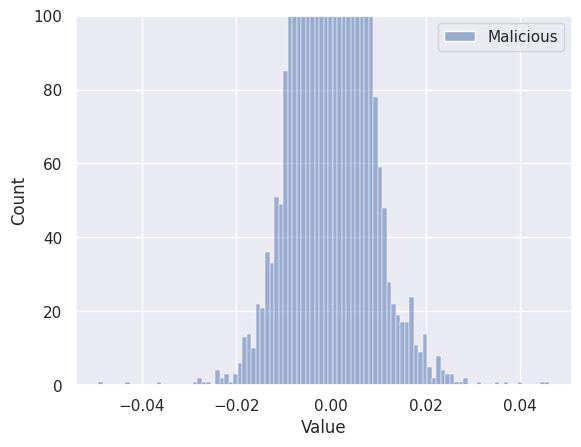}
    \caption{Gradient element value histogram (zoom-in)}
  \end{subfigure}
  \caption{Gradient element value histogram of a malicious gradient.}
  \label{figure: gradient element value histogram}
\end{figure}

We further empirically verify our analysis in Section \ref{section: Backdoor attacks over flat loss landscape}, showing that mimicking benign clients has a low penalty. In this experiment, we consider two gradient vectors, one from a benign client and the other one from a malicious client. Figure \ref{figure: gradient element value histogram} plots the histogram of malicious gradient elements and suggests that a majority of them are small, indicating a flat loss landscape. For example, there are only 747 out of 11689512 gradient elements whose absolute value is greater than 0.01. Then, we let the malicious gradient mimic a benign gradient by setting the value of all malicious gradient elements that are not among the 1000 largest ones to the corresponding benign gradient elements. Such a mimicking strategy increases the cosine similarity between the benign and the malicious gradient to 0.983 while only slowing down the \emph{increase} of the attack success loss by 3.85\%.

\subsection{Separable Minima}
% We employ 5 federated learning clients with benign data samples and train 5 local federated learning models (Resnet-18) over the clients. Then, we randomly generate 1000 convex combinations of the 5 local models and evaluate the accuracy and attack success rate (ASR). The ASR is always 0\% % across all convex combinations, although the accuracy can decrease. Such a result suggests that parameters within the convex hull of benign minima are unlikely to have increased ASR and are consistent with the separable minima assumption (Assumption \ref{assumption: separable minimum}).
We employ 5 federated learning clients with benign data samples and train a global federated learning model. Then, we let each client do one round of local fine-tuning (a.k.a. personalization) and find that $\bm{w}' \notin \mathcal{W}^*$.

\subsection{Ablation Study}
\label{appendix: ablation}
We include ablation studies to demonstrate that the AND-mask and trimmed-mean estimator are necessary for defending against backdoor attacks. Table \ref{table: ablation} summarizes the results and shows that neither AND-mask nor trimmed-mean estimator defends against backdoor attacks alone. We also replace the sign consistency in AND-mask using the sample mean-variance ratio.
% For the robustness degradation of using the sample mean-variance ratio, we hypothesize that hurting benign features can exacerbate the backdoor attack because the trigger has fewer benign features to compete.

\begin{table*}[h!]
\centering
\caption{Accuracy of Aggregators Under Edge-case Backdoor Attack.}
\begin{threeparttable}
\begin{tabular}{l|@{\hspace{.5\tabcolsep}}c@{\hspace{.5\tabcolsep}}c|@{\hspace{.5\tabcolsep}}c@{\hspace{.5\tabcolsep}}c|@{\hspace{.5\tabcolsep}}c@{\hspace{.5\tabcolsep}}c|@{\hspace{.5\tabcolsep}}c@{\hspace{.5\tabcolsep}}c}
\toprule
\multirow{2}{*}{Method} &
\multicolumn{2}{c|@{\hspace{.5\tabcolsep}}}{CIFAR-10} &
\multicolumn{2}{c|@{\hspace{.5\tabcolsep}}}{Twitter} &
\multicolumn{2}{c|@{\hspace{.5\tabcolsep}}}{Phishing} 
\\
& Acc & ASR & Acc & ASR & Acc & ASR \\
\midrule
Ours & .677 {\tiny $\pm$ .001} & .001 {\tiny $\pm$ .001} & .687 {\tiny $\pm$ .001} & .296 {\tiny $\pm$ .003} & .999 {\tiny $\pm$ .001} & .000 {\tiny $\pm$ .001} \\
AND-mask & .672 {\tiny $\pm$ .001} & .655 {\tiny $\pm$ .001} & .652 {\tiny $\pm$ .001} & .493 {\tiny $\pm$ .017} & .999 {\tiny $\pm$ .001} & .999 {\tiny $\pm$ .001} \\
Trimmed-mean & .687 {\tiny $\pm$ .001} & .512 {\tiny $\pm$ .001} & .728 {\tiny $\pm$ .001} & .640 {\tiny $\pm$ .016} & .999 {\tiny $\pm$ .001} & .999 {\tiny $\pm$ .001} \\
Mean-Variance Ratio & .554 {\tiny $\pm$ .001} & .000 {\tiny $\pm$ .001} & .613 {\tiny $\pm$ .001} & .603 {\tiny $\pm$ .001} & .999 {\tiny $\pm$ .000} & .333 {\tiny $\pm$ .333} \\
\bottomrule
\multicolumn{7}{c}{\footnotesize Note: The numbers are average accuracy over three runs. Variance is rounded up.}
\end{tabular}
\end{threeparttable}
\label{table: ablation}
\end{table*}

\subsection{Hyper-parameter Sensitivity}
\label{appendix: sensitivity}
\begin{figure*}[h!]
  \centering
  \begin{subfigure}[c]{0.325\linewidth}
    \centering\includegraphics[width=\textwidth]{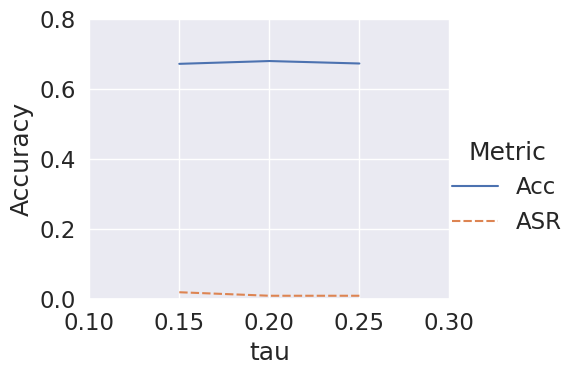}
    \caption{Varying $\tau$}
  \end{subfigure}
  \begin{subfigure}[c]{0.325\linewidth}
    \centering\includegraphics[width=\textwidth]{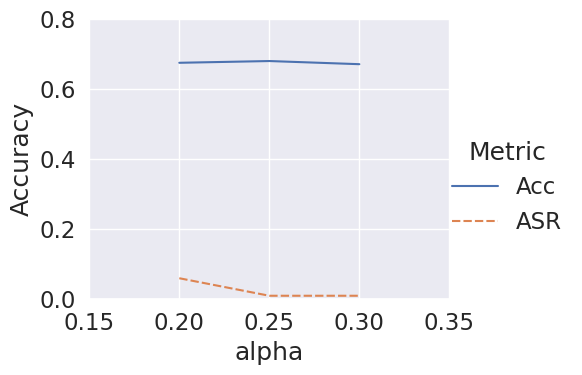}
    \caption{Varying $\alpha$}
  \end{subfigure}
  \begin{subfigure}[c]{0.325\linewidth}
    \centering\includegraphics[width=\textwidth]{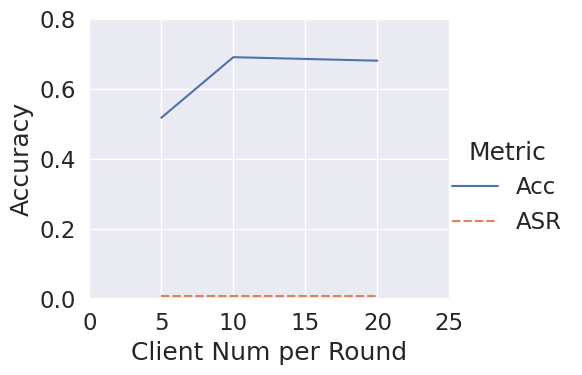}
    \caption{Varying client number}
  \end{subfigure}
  \caption{Performance of our defense with varying hyper-parameters.}
  \label{figure: hyper-param sensitivity}
\end{figure*}

This section provides additional experiments using the CIFAR-10 dataset to show that our approach is easy to apply and does not require expensive hyperparameter tuning. Our experiments measure the impact of the masking threshold $\tau$ and trim level $\alpha$ on our defense and evaluate our approach with fewer clients per round. Experimental results in Figure \ref{figure: hyper-param sensitivity} suggest that there exists a wide range of threshold configurations that are effective against backdoor attacks. Moreover, our approach can work with as few as 10 clients per round.

\subsection{Result Discussion}
\textbf{Vector-wise.} Common vector-wise defenses such as Krum estimate pair-wise similarities in terms of Euclidean distance \citep{Blanchard2017MachineLW} (Krum and multi-Krum) of cosine similarity \citep{Nguyen2021FLAMETB} (multi-Krum$_C$) between each gradient and others. The gradients that are dissimilar to others are removed. The vector-wise view is insufficient for defending against backdoor attacks because backdoor attacks can succeed by manipulating a tiny subset of parameters (e.g. 5\%) \citep{wu2021adversarial} without incurring much vector-wise difference. In practice, we observe that the malicious gradients can get high similarity scores and circumvent vector-wise defenses. We also find that Krum achieves much lower accuracies than the results from previous works \citep{NEURIPS2020_b8ffa41d}. We attribute such a difference to the capability of adversaries. We employ a strong adversary that can compromise 20\% clients \citep{karimireddy2022byzantinerobust} at every round, while prior work \citep{NEURIPS2020_b8ffa41d} lets an adversary compromise 1 client every 10 rounds.

% We adopt (Multi-)Krum \citep{} as a base for vector-wise defense. Krum estimates pair-wise similarities between each gradient and others. The gradients that are dissimilar to others are removed. Krum uses Euclidean distance to estimate the similarity. We also include an improved multi-Krum$_C$ that uses cosine difference to estimate the similarity. Server recent works \citep{} suggest that that cosine difference can help detect the malicious clients that perform backdoor attacks.

\textbf{Dimension-wise.}
Figure \ref{figure: median} in Section \ref{section: motivating setting} shows the limitation of the trimmed-mean estimator, which was the most effective defense against the edge-case backdoor attack. We also include Sign-SGD with majority vote \citep{Bernstein2019signSGDWM} as a defense, which binarizes the gradient and takes the majority vote as the aggregation result. However, Sign-SGD struggles to train a large federated model (e.g., Resnet-18 on CIFAR-10) and can suffer from the same failure mode as the median estimator where the clients have diverse signs. Then, the adversary can put more weight on one side and mislead the voting result.

\textbf{Combination.} A naive combination of multi-Krum and the trimmed-mean estimator fails to defend against the backdoor attack because neither multi-Krum nor the trimmed-mean estimator avoids the failure mode of the other.

\textbf{Weak-DP.} The weak-DP defense \citep{Sun2019CanYR} first bounds the gradient norms, then adds additive noise (e.g., Gaussian noise) to the gradient vector. The edge-case backdoor attack can work without scaling up the gradients, circumventing the norm bounding. For the additive noise, we hypothesize that in some dimensions, the difference between malicious and benign gradients can be too large for the Gaussian noise to blur their boundary.

% \textbf{Freezing Layers.} 
% Since we employ a pre-trained Resnet-18 \citep{He2016DeepRL} on CIFAR-10, freezing the convolution layers may avoid entangling the trigger. However, this approach lacks empirical robustness, possibly because the adversary can use semantic features (e.g., blue color on airplanes \citep{NEURIPS2020_b8ffa41d}) that the pre-trained model already learns as triggers. 

\textbf{Advanced Defenses.} RFA \citep{9721118} computes geometric medians as the aggregation result, which is shown to be ineffective \citep{NEURIPS2020_b8ffa41d}. SparseFed \citep{pmlr-v151-panda22a} only accepts elements with large magnitude in the aggregation results. However, both benign and malicious updates can contribute to large magnitudes. The limitations of FLTrust and trigger inversion are discussed in Section \ref{section: motivating setting}.

\textbf{Robust Learning Rate (RLR).} The RLR \citep{ozdayi2021defending} approach is similar to ours, but its design lacks a robustness guarantee. The RLR approach flips the gradient sign if a dimension’s sign consistency is low. Although such a strategy can recover the correct gradient direction if the aggregation result is misled and has the same sign as the malicious gradient, it can proactively mislead the aggregation result if the adversary fails. Suppose the adversary uploads a gradient -0.1 in a low-consistency dimension but does not successfully mislead the aggregation result (e.g., 0.5). Then, using the RLR strategy leads to an aggregation result –0.5 and helps the adversary. In contrast, our approach directly sets the aggregation result to 0 in dimensions with low consistency and applies a trimmed-mean estimator for other dimensions.

\begin{table*}[h]
\centering
\caption{Accuracy of Invariant Aggregators Under Various Continuous Attack.}
\begin{threeparttable}
\begin{tabular}{l|@{\hspace{.5\tabcolsep}}c@{\hspace{.5\tabcolsep}}c|@{\hspace{.5\tabcolsep}}c@{\hspace{.5\tabcolsep}}c|@{\hspace{.5\tabcolsep}}c@{\hspace{.5\tabcolsep}}c|@{\hspace{.5\tabcolsep}}c@{\hspace{.5\tabcolsep}}c}
\toprule
\multirow{2}{*}{Method} &
\multicolumn{2}{c|@{\hspace{.5\tabcolsep}}}{CIFAR-10} &
\multicolumn{2}{c|@{\hspace{.5\tabcolsep}}}{Twitter} &
\multicolumn{2}{c|@{\hspace{.5\tabcolsep}}}{Phishing} 
\\
& Acc & ASR & Acc & ASR & Acc & ASR \\
\midrule
No Attack & .685 {\tiny $\pm$ .001} & .000 {\tiny $\pm$ .001} & .691 {\tiny $\pm$ .001} & .095 {\tiny $\pm$ .001} & .999 {\tiny $\pm$ .001} & .000 {\tiny $\pm$ .001} \\
\hline
Edge-case \citep{NEURIPS2020_b8ffa41d} & .677 {\tiny $\pm$ .001} & .001 {\tiny $\pm$ .001} & .687 {\tiny $\pm$ .001} & .296 {\tiny $\pm$ .003} & .999 {\tiny $\pm$ .001} & .000 {\tiny $\pm$ .001} \\
DBA \citep{Xie2020DBA} & .679 {\tiny $\pm$ .001} & .001 {\tiny $\pm$ .001} & N\textbackslash A & N\textbackslash A & .999 {\tiny $\pm$ .001} & .000 {\tiny $\pm$ .001} \\
\hline
Edge+Collude \citep{pmlr-v151-panda22a} & .675 {\tiny $\pm$ .001} & .021 {\tiny $\pm$ .001} & .685 {\tiny $\pm$ .002} & .324 {\tiny $\pm$ .001} & .999 {\tiny $\pm$ .001} & .000 {\tiny $\pm$ .001} \\
Edge+Adaptive & .668 {\tiny $\pm$ .001} & .013 {\tiny $\pm$ .001} & .682 {\tiny $\pm$ .001} & .331 {\tiny $\pm$ .002} & .999 {\tiny $\pm$ .001} & .000 {\tiny $\pm$ .001} \\
Edge+Collude+Adaptive & .667 {\tiny $\pm$ .001} & .049 {\tiny $\pm$ .001} & .683 {\tiny $\pm$ .001} & .335 {\tiny $\pm$ .001} & .999 {\tiny $\pm$ .001} & .000 {\tiny $\pm$ .001} \\
\bottomrule
\multicolumn{7}{c}{\footnotesize Note: The numbers are average accuracy over three runs. Variance is rounded up.}
\vspace{-0.1in}
\end{tabular}
\end{threeparttable}
\label{table: more attacks}
\end{table*}

\subsection{Results under Various Backdoor Attacks}

\paragraph{Strategies.} We extensively evaluate our approach against multiple state-of-the-art attack strategies (Table \ref{table: more attacks}), including distributed backdoor attack (DBA) \citep{Xie2020DBA}, colluding attack \citep{pmlr-v151-panda22a}, and an adaptive attack that aims to leverage the unmasked gradient elements (Appendix \ref{appendix: experimental setup}). Our invariant aggregator remains effective with minor ASR increase under all strategies and their compositions. We further compare our approach against strong baselines under the collude and adaptive strategies on the CIFAR-10 dataset(Tables \ref{table: collude adversary} and \ref{table: adaptive adversary}).

\begin{table*}[h!]
  \caption{Performance of Aggregators Under Collude Adversary on the CIFAR-10 Dataset.}
  \centering
  \begin{tabular}{l|c|c|c|c}
    \toprule
    Collude Adversary & Ours & FLTrust & RFA & FLIP \\
    \midrule
    Acc & .675 {\tiny $\pm$ .002} & .671 {\tiny $\pm$ .001} & .681 {\tiny $\pm$ .002} & .665 {\tiny $\pm$ .001} \\
    ASR & .021 {\tiny $\pm$ .001} & .580 {\tiny $\pm$ .003} & .878 {\tiny $\pm$ .002} & .271 {\tiny $\pm$ .001} \\
    \bottomrule
  \end{tabular}
  \label{table: collude adversary}
\end{table*}

\begin{table*}[h!]
  \caption{Performance of Aggregators Under Adaptive Adversary on the CIFAR-10 Dataset.}
  \centering
  \begin{tabular}{l|c|c|c|c}
    \toprule
    Adaptive Adversary & Ours & FLTrust & RFA & FLIP \\
    \midrule
    Acc & .668 {\tiny $\pm$ .001} & .672 {\tiny $\pm$ .002} & .683 {\tiny $\pm$ .001} & .668 {\tiny $\pm$ .001} \\
    ASR & .013 {\tiny $\pm$ .001} & .543 {\tiny $\pm$ .003} & .837 {\tiny $\pm$ .004} & .236 {\tiny $\pm$ .001} \\
    \bottomrule
  \end{tabular}
  \label{table: adaptive adversary}
\end{table*}

\paragraph{Poison Ratio.} 
We evaluate our approach with various poison ratios (Table \ref{table: poison ratio}), which means the percentage of malicious clients in a federated learning system.

\begin{table*}[h!]
  \caption{Performance of Invariant Aggregator under Various Poison Ratio on the CIFAR-10 Dataset.}
  \centering
  \begin{tabular}{l|c|c|c|c|c}
    \toprule
    Poison Ratio & 0\% & 5\% & 10\% & 15\% & 20\% \\
    \midrule
    Acc & .687 {\tiny $\pm$ .002} & .683 {\tiny $\pm$ .003} & .681 {\tiny $\pm$ .001} & .681 {\tiny $\pm$ .002} & .677 {\tiny $\pm$ .001} \\
    ASR & .000 {\tiny $\pm$ .001} & .000 {\tiny $\pm$ .001} & .000 {\tiny $\pm$ .001} & .001 {\tiny $\pm$ .001} & .001 {\tiny $\pm$ .001} \\
    \bottomrule
  \end{tabular}
  \label{table: poison ratio}
\end{table*}

\paragraph{Trigger Size.}
We evaluate our approach with various pixel trigger sizes on the CIFAR-10 dataset (Table \ref{table: trigger size}).

\begin{table*}[h!]
  \caption{Performance of Invariant Aggregator under Various Trigger Sizes on the CIFAR-10 Dataset.}
  \centering
  \begin{tabular}{l|c|c|c}
    \toprule
    Trigger Size & $\times 1$ & $\times 2$ & $\times 4$ \\
    \midrule
    Acc & .679 {\tiny $\pm$ .001} & .677 {\tiny $\pm$ .001} & .675 {\tiny $\pm$ .002} \\
    ASR & .001 {\tiny $\pm$ .001} & .003 {\tiny $\pm$ .001} & .003 {\tiny $\pm$ .001} \\
    \bottomrule
  \end{tabular}
  \label{table: trigger size}
\end{table*}
\subsection{Results under Different Data Heterogeneity Level.}
We evaluate our approach under different data heterogeneity levels on the CIFAR-10 dataset (Table \ref{table: heterogeneity level}). The parameter $\alpha$ in the Dirichlet distribution-based non-i.i.d. data partition controls the heterogeneity level.

\begin{table*}[h!]
  \caption{Performance of Invariant Aggregator under Various Data Heterogeneity Level on the CIFAR-10 Dataset. The parameter $\alpha$ in the Dirichlet distribution-based non-i.i.d. data partition controls the heterogeneity level. The i.i.d.(full) means every client has access to the full dataset locally.}
  \centering
  \begin{tabular}{l|c|c|c|c}
    \toprule
    Heterogeneity & i.i.d. (full) & i.i.d. ($\alpha = 1.0$) & non-i.i.d. ($\alpha = 0.5$) & non-i.i.d. ($\alpha = 0.2$) \\
    \midrule
    Acc & .857 {\tiny $\pm$ .003} & .696 {\tiny $\pm$ .002} & .677 {\tiny $\pm$ .003} & .632 {\tiny $\pm$ .001} \\
    ASR & .000 {\tiny $\pm$ .001} & .000 {\tiny $\pm$ .001} & .001 {\tiny $\pm$ .001} & .001 {\tiny $\pm$ .001} \\
    \bottomrule
  \end{tabular}
  \label{table: heterogeneity level}
\end{table*}

\end{document}